\documentclass[draft]{agujournal2019}
\usepackage{amsmath,amssymb}
\usepackage{booktabs}
\draftfalse

\journalname{ }

\newcommand{\Q}{\mathbf{Q}}

\newcommand{\nvec}{\mathbf{n}}
\newcommand{\x}{\mathbf{x}}
\newcommand{\R}{\mathbb{R}}
\newcommand{\norm}[1]{\left\|#1\right\|}

\begin{document}

\title{Finite Volume-Informed Neural Network Framework for 2D Shallow Water Equations: Rugged Loss Landscapes and the Importance of Data Guidance}

\authors{Xiaofeng Liu\affil{1,2}}

\affiliation{1}{Department of Civil and Environmental Engineering, Pennsylvania State University, University Park, PA, USA}
\affiliation{2}{Institute for Computational and Data Sciences, Pennsylvania State University, University Park, PA, USA}

\correspondingauthor{Xiaofeng Liu}{xzl123@psu.edu}

\begin{keypoints}
\item A differentiable well-balanced Roe finite-volume loss is introduced for 2D shallow-water PINNs on unstructured meshes
\item The loss landscape of physics-only FVM-PINN is rugged with many shallow minima which nearly satisfy PDEs
\item Data guidance, even with sparse measurements, breaks the degeneracy and is essential for convergence to the correct solution
\end{keypoints}

\begin{abstract}
Physics-informed neural networks (PINNs) are a simple surrogate-modelling paradigm for partial differential equations, but their standard strong-form residual formulation is ill suited to the shallow water equations (SWE). It cannot enforce local conservation, handle discontinuities, or leverage the boundary-conforming unstructured meshes used in real-world applications. We introduce ``Data-Guided FVM-PINN'', a framework that replaces the strong-form residual with a differentiable, well-balanced Roe Riemann-solver finite-volume (FVM) loss evaluated on unstructured meshes. The major finding is that physics-only FVM-PINN training often fails on realistic 2D problems: the network collapses to a trivial low-momentum state that nearly satisfies the FVM-PINN residual but bears no resemblance to the true flow. A loss-landscape diagnostic shows that the FVM-PINN loss at zero momentum is only about $7\times$ larger than at the trained solution, a shallow basin that an ordinary optimizer falls into; adding even sparse data turns this into a $310\times$ separation, breaking the degeneracy. On a 2D block-in-channel benchmark, just $200$ random velocity measurements drop the velocity-field $L_2$ error by $22\times$ versus physics-only; $50$ measurements still deliver a $7\times$ reduction. A controlled ablation isolates the contribution of the FVM-PINN loss: it reduces velocity-field $L_2$ by $\sim$$23\%$ in the sparse-data regime and is essentially neutral when dense reference data is available. On a real-world Savannah River reach ($1306$ cells, $3600$~s simulation, five Manning zones), the framework constructs an accurate surrogate from SRH-2D anchor data, with time-window decomposition reducing error monotonically via progressive initial-condition handoff. 
\end{abstract}

\section*{Plain Language Summary}
Floods, storm surges, tsunami inundation, lake and reservoir circulation, and other free-surface water flows across the geosciences are routinely simulated by numerical solvers that work on unstructured meshes. While accurate, these solvers must be re-run from scratch for every new scenario and are difficult to combine with field measurements. Neural-network surrogates promise much faster predictions and built-in sensitivity analysis, but the standard physics-informed neural network does not respect mass and momentum conservation and is poorly suited to the kind of meshes used for real-world applications. We replace the physics constraint of the neural network with a finite-volume balance evaluated on real meshes from the popular SRH-2D model. Our central finding is that the physics constraint alone is not enough: the network can collapse to a trivial near-zero-flow state that nominally satisfies the equations but does not match reality, and even a small number of velocity measurements is sufficient to break this degeneracy and pin the network to the correct solution. For long-time simulations of a real river reach, splitting the time interval into shorter windows that hand off initial conditions to one another further improves accuracy. The same methodology is directly applicable to other depth-averaged flow problems such as coastal storm surge, tsunami inundation, and lake or reservoir circulation.

\section{Introduction}
\label{sec:intro}

The shallow water equations (SWEs) are a backbone mathematical model across water resources and geophysical fluid dynamics. They describe free-surface flow in rivers, estuaries, coastal areas, lakes, and urban flood plains; and geophysical hazards such as glacial-lake-outburst floods and dam-break waves. The SWEs, in their various simplified forms, also underpin the surface-water routing schemes embedded in hydrologic land-surface models and Earth-system models, where they provide the dynamical bridge between continental hydrology, coastal oceans, and the climate system. Operational forecasting, climate-impact assessment, and risk mapping across these domains all rely on the SWEs as a workhorse model, and improvements in how the SWEs are solved propagate directly to operational practice across many fields.

Numerical models based on the finite volume method (FVM), such as HEC-RAS, SRH-2D, and LISFLOOD \cite{brunner1995hec,lai2010srh2d, lai2008srh2d_manual,van2010lisflood}, provide accurate solutions on arbitrary unstructured meshes and are the standard tools in fluvial hydraulics and coastal flows for flood-hazard assessment, infrastructure design, floodplain mapping, and coastal-inundation studies. Despite their accuracy, FVM solvers have fundamental limitations from a machine-learning and inverse-problem perspective. Each new set of boundary conditions, initial conditions, or parameter values requires an independent simulation. Analytic gradients of the solution with respect to inputs and parameters are not easily available. And real-time inference is infeasible for operational forecasting and ensemble-scale uncertainty quantification.

These limitations place the SWEs at the centre of the recent push toward physics-aware machine learning and differentiable computation for flow modelling \cite{liu2025uswe}. Physics-informed and operator-learning surrogates promise gradient-bearing, fast-to-evaluate alternatives to traditional hyperbolic solvers, with direct applications to inverse problems, sensitivity analysis, parameter calibration, and data assimilation across the geophysical domains listed above. Realizing this promise on realistic application configurations, which often include non-trivial bathymetry, multiple roughness regions, and shock-like features, requires a discretisation-faithful coupling of conservation-law structure with neural surrogates on the unstructured meshes used by operational hydrological and hydraulics models. This direction is part of a broader move toward differentiable Earth-system components, in which each block of a coupled hydrologic-coastal-climate pipeline becomes gradient-aware and amenable to ensemble data assimilation, inverse problems, and uncertainty quantification at scale \cite{Shen2023}. The 2D SWEs on unstructured meshes is one of the first such components for which a fully differentiable, well-balanced, shock-capturing neural surrogate is within reach \cite{liu2025uswe}.

Among many approaches for constructing neural surrogate, physics-informed neural networks (PINNs) \cite{raissi2019pinn} are perhaps the simplest one. This work uses PINN. However, the methodology can be extended to other approaches such as neural operator learning with moderate ease. PINN offers a complementary capability: a neural network parameterises the solution $\Q(\x, t)$ and is trained by minimising a loss that encodes the partial differential equations (PDEs), boundary/initial conditions, and available observations. Once trained, the network evaluates at arbitrary space-time points in milliseconds, provides analytic gradients, and can be differentiated with respect to unknown physical parameters for inversion. However, standard PINNs have well-known weaknesses that become critical for the SWEs: (i) the strong-form residual assumes smooth solutions, but SWE solutions routinely contain discontinuities; (ii) minimising a pointwise residual provides no guarantee that mass or momentum is locally and globally conserved; (iii) there is no mechanism to enforce the entropy condition that distinguishes physical shocks from non-physical ones; and (iv) PINNs operate on scattered collocation points with no notion of mesh connectivity, making it difficult to leverage the boundary-conforming unstructured meshes used in practice.

In the literature, a first wave of work has applied PINNs to the SWE using the strong-form PDE residual on collocation points. \citeA{bihlo2022pinn_swe_sphere} solved the SWE on the sphere with a multi-model time-windowing approach, and \citeA{dazzi2024pinn_swe} treated the SWE with topography while discussing data assimilation as a secondary feature. \citeA{brecht2025pinn_tsunami} extended PINNs to tsunami inundation modelling, and \citeA{tian2025pinn_swe} addressed the 2D SWE with rainfall and terrain source terms. \citeA{qian2024pinn_swe_datafree} demonstrated data-free PINNs for the 2D SWE on simple analytical benchmarks. None of these works addresses cell-wise conservation, shock capturing through Riemann solvers, or boundary-conforming unstructured meshes, which are the three properties that together define the real-world application setting we target.

A second line of work has begun to address these limitations by replacing the strong-form PDE residual with finite-volume- or Godunov-based losses. \citeA{wei2025fvmpinn} and \citeA{wei2025ffvpinn} replaced automatic differentiation with FVM discretisation for incompressible flows on structured grids, achieving substantial speedups. \citeA{cassia2025godunov_loss} introduced Godunov loss functions using the HLLC Riemann solver for the 2D compressible Euler equations, demonstrating shock-capturing ability unavailable from strong-form PINNs. \citeA{su2024fvpinn} used Gauss's theorem to reduce the order of the required derivatives, \citeA{mei2024ufvpinn} embedded a differentiable FVM solver for heterogeneous PDEs, and \citeA{zhu2026sfvnet} developed a U-Net with an FVM loss for the compressible Euler system. For 1D problems, \citeA{urban2025lrpinn} incorporated Roe's Riemann solver for hydrodynamics, \citeA{oubarka2026wepinn} proposed weak-form and entropy-aware PINNs on mesh-free control volumes, and \citeA{patsatzis2025gorinns} introduced GoRINNs, which combine shallow neural networks with Godunov-type FVM schemes to learn unknown flux closures, demonstrated on 1D systems including the SWE. All of these methods either operate on structured Cartesian grids, are mesh-free, or are restricted to 1D, and none targets the 2D SWEs on unstructured meshes, which is the standard configuration in real-world applications.

A complementary literature documents failure modes of standard PINNs for stiff or hyperbolic systems and proposes alternative formulations. \citeA{krishnapriyan2021failure} characterised optimisation failures of strong-form PINNs and proposed time-marching curricula as a remedy. For hyperbolic conservation laws specifically, \citeA{deryck2024wpinn} introduced \emph{wPINNs} - weak PINNs that penalise a dual-norm residual - and proved they approximate the entropy solution; their construction is mesh-free and tested on Burgers' equation. \citeA{jagtap2020cpinn} introduced conservative PINNs (cPINNs) on discrete domains, and \citeA{patel2022conservative} proposed thermodynamically consistent space-time control-volume PINNs for hyperbolic systems, enforcing conservation in ways conceptually related to but distinct from our face-based Roe formulation. The failure mode we document in this work is complementary to these: rather than incorrect wave speeds, we identify a trivial low-momentum minimum of the FVM residual that the optimizer falls into in the absence of supervisory data, which is a structural property of any cell-conservation loss whose face flux contributions vanish at zero velocity.

The original PINN formulation \cite{raissi2019pinn} supports observational data as an additional loss term, primarily as a feature for inverse problems. For forward problems on simple geometries, data is conventionally treated as optional. For FVM-PINN on the realistic 2D SWEs, we instead show that data is essential: it is the mechanism that breaks the trivial-minimum degeneracy and guides the optimizer out of shallow basins in the loss landscape. We additionally quantify the contribution of the FVM residual itself across data densities through physics-on/off ablations. This contribution-decomposition methodology, to our knowledge, has not been applied to FVM-PINN training before.

To address the limitations described above, the contributions of this work are as follows. We propose Data-Guided FVM-PINN, a framework that retains the conservation-enforcing properties of an FVM-based loss while incorporating observational or simulation data for robust convergence on realistic problems. Specifically, (i) we construct a differentiable, well-balanced Roe approximate Riemann solver of \citeA{rogers2003mathematical} and \citeA{liu2025uswe}, working directly on the perturbation-form state $\Q = [\xi, uh, vh]$; unlike all prior FVM-PINN methods \cite{wei2025fvmpinn, wei2025ffvpinn, su2024fvpinn, mei2024ufvpinn, zhu2026sfvnet, cassia2025godunov_loss}, which operate on structured Cartesian grids and target either incompressible flows or compressible Euler systems, our framework natively supports unstructured meshes; (ii) we document and quantitatively diagnose a physics-only failure mode in which the FVM residual loss admits a near-trivial low-momentum minimum that the optimizer easily falls into, producing solutions that nearly satisfy conservation but bear no resemblance to the true flow; a loss-landscape diagnostic (Section~\ref{sec:loss_landscape}) shows the FVM basin at zero momentum is only $7\times$ deeper than at the trained solution while the data loss separates the two states; (iii) we show that sparse observational data is sufficient to break the trivial-minimum degeneracy in the 2D SWEs; and (iv) we apply the framework to a $1$~km Savannah River reach using SRH-2D anchor data as supervision, where time-window decomposition reduces error monotonically through progressive IC handoff.

The remainder of the paper is organised as follows. Section~\ref{sec:background} describes the 2D SWEs in perturbation form, the FVM discretisation, and the standard PINN formulation. Section~\ref{sec:method} presents the FVM-PINN formulation including data-guided training. Section~\ref{sec:scalability} describes the scalability strategies. Section~\ref{sec:experiments} reports the numerical experiments. Section~\ref{sec:discussion} discusses the loss-landscape diagnostic and the regime-dependent contribution of the FVM residual. Section~\ref{sec:limitations} reflects on the limitations of the present work and outlines future directions. Section~\ref{sec:conclusions} concludes.

\section{Governing Equations, FVM Discretisation, and PINN Formulation}
\label{sec:background}

\subsection{The 2D Shallow Water Equations in Perturbation Form}
\label{sec:swe}

\begin{figure}[h!]
  \centering
  \includegraphics[width=0.9\textwidth]{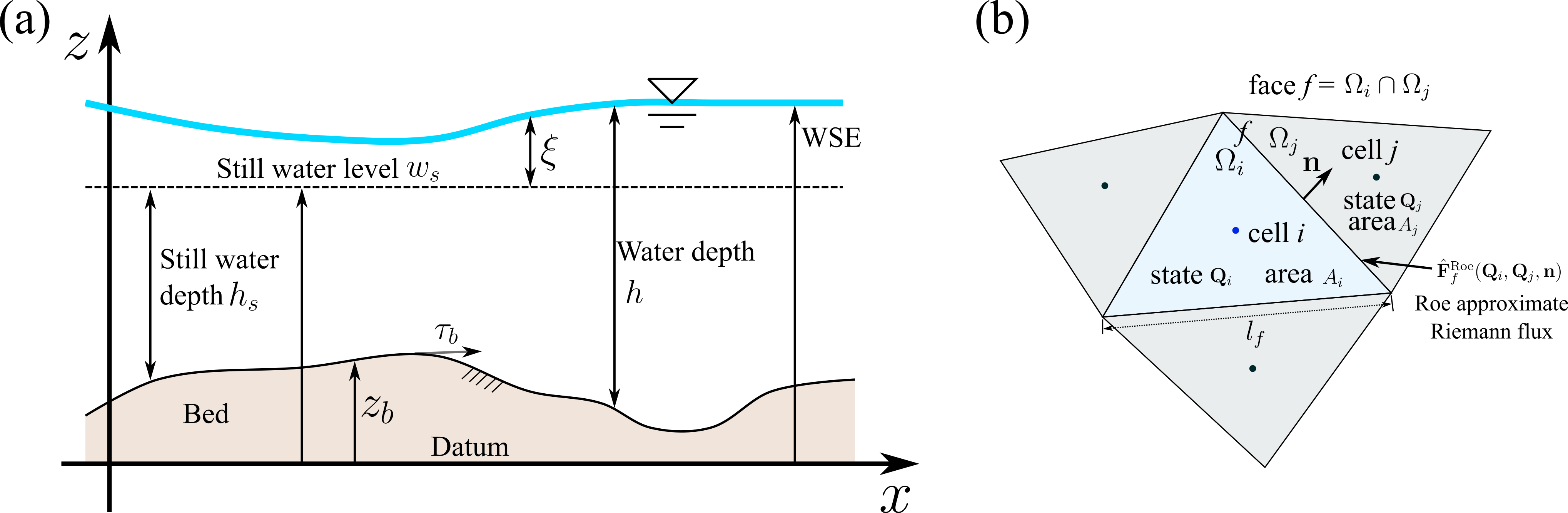}
  \caption{Schematic of the 2D shallow water equations: (a) the physical domain and definitions, and (b) the unstructured finite-volume mesh with face-based fluxes.}
  \label{fig:swe_scheme_fvm}
\end{figure}

We adopt the well-balanced perturbation form of the 2D SWEs introduced by \citeA{rogers2003mathematical} and used in the universal-SWE differentiable solver of \citeA{liu2025uswe}. The schematic of the 2D SWEs in perturbation form is shown in Figure~\ref{fig:swe_scheme_fvm}(a). The conserved variables are
\begin{equation}
  \Q = \begin{pmatrix} \xi \\ uh \\ vh \end{pmatrix},
  \qquad
  \xi(\x,t) = h(\x,t) - h_s(\x),
\label{eq:swe_q}
\end{equation}
where $h$ is the water depth, $h_s = \max(0, w_s - z_b)$ is the still-water depth above the bed elevation $z_b$ for a chosen reference water-surface elevation $w_s$, and $\xi$ is the free-surface perturbation. $\x = (x, y)$ denotes the spatial coordinates. The benefit of equation~\eqref{eq:swe_q} over the standard $[h, hu, hv]$ form is that $\xi \approx 0$ at lake-at-rest, making the well-balanced property an algebraic identity rather than a numerical  near-cancellation.

In conservation form, the SWEs read
\begin{equation}
  \frac{\partial \Q}{\partial t}
  + \frac{\partial \mathbf{F}(\Q)}{\partial x}
  + \frac{\partial \mathbf{G}(\Q)}{\partial y}
  = \mathbf{S}(\Q),
\label{eq:swe_strong}
\end{equation}
with fluxes and source term \cite{liu2025uswe}
\begin{equation}
  \mathbf{F} = \begin{pmatrix}
    uh \\
    u^2 h + \tfrac{1}{2} g (\xi^2 + 2\xi h_s) \\
    uvh
  \end{pmatrix},
  \quad
  \mathbf{G} = \begin{pmatrix}
    vh \\
    uvh \\
    v^2 h + \tfrac{1}{2} g (\xi^2 + 2\xi h_s)
  \end{pmatrix},
  \quad
  \mathbf{S} = \begin{pmatrix}
    0 \\
    -\tau_{bx}/\rho - g\,\xi\,S_{0x} \\
    -\tau_{by}/\rho - g\,\xi\,S_{0y}
  \end{pmatrix},
\label{eq:swe_vars}
\end{equation}
where $g$ is gravitational acceleration, $\rho$ is the water density, $S_{0x,y} = - \partial z_b / \partial x,y$ are bed slopes, and the
bed shear stress is parameterised through Darcy--Weisbach friction 
\begin{equation}
  (\tau_{bx},\, \tau_{by}) = \tfrac{\rho}{8} f \sqrt{u^2 + v^2}\,(u, v),
  \qquad
  f = \frac{8 g n^2}{h^{1/3}},
\label{eq:friction}
\end{equation}
with $n$ the Manning roughness coefficient. The magnitude of the velocity vector $\mathbf{u} = (u, v)$ is $\sqrt{u^2 + v^2}$.

The reformulated pressure flux $\tfrac{1}{2} g (\xi^2 + 2\xi h_s)$
in equation~\eqref{eq:swe_vars} arises from the source-term split
\begin{equation}
  g h \frac{\partial \xi}{\partial x}
  = \frac{\partial}{\partial x}\!\left[\tfrac{1}{2} g (\xi^2 + 2\xi h_s)\right]
  - g h \frac{\partial z_b}{\partial x}
\label{eq:source_split}
\end{equation}
introduced by \citeA{rogers2003mathematical}.
Equation~\eqref{eq:source_split} guarantees that the Roe flux of
Section~\ref{sec:fvm} balances the bed-slope source exactly when
$\xi = 0$ and $u = v = 0$, so a quiescent state remains undisturbed under
arbitrary topography.
The system is hyperbolic with eigenvalues
$\lambda_{1,2,3} = u_n - c,\, u_n,\, u_n + c$,
$u_n = \mathbf{u}\cdot\nvec$, $c = \sqrt{g h}$.

Integrating equation~\eqref{eq:swe_strong} over a control volume $\Omega_i$ and applying the divergence theorem yields the integral law
\begin{equation}
  \frac{d}{dt}\int_{\Omega_i} \Q \, dA
  + \oint_{\partial \Omega_i} \bigl[\mathbf{F}(\Q),\, \mathbf{G}(\Q)\bigr]\cdot \nvec \, d\ell
  = \int_{\Omega_i} \mathbf{S}(\Q) \, dA,
\label{eq:integral_law}
\end{equation}
which holds even when $\Q$ is discontinuous on $\partial\Omega_i$. Equation~\eqref{eq:integral_law} is the foundation of FVM-PINN loss developed in Section~\ref{sec:method}.

\subsection{Finite Volume Discretisation}
\label{sec:fvm}

Equation~\eqref{eq:integral_law} can be discretised on an unstructured mesh (Figure~\ref{fig:swe_scheme_fvm}(b)). Approximating the cell-average $\bar{\Q}_i \approx \Q_i$ and face integrals by midpoint quadrature gives the semi-discrete system 
\begin{equation}
  A_i \frac{d \bar{\Q}_i}{dt}
  + \sum_{f \in \partial \Omega_i} \hat{\mathbf{F}}_f \, \ell_f
  = A_i \bar{\mathbf{S}}_i,
\label{eq:fvm_discrete}
\end{equation}
where $A_i$ is the cell area, $\ell_f$ the face length, and $\hat{\mathbf{F}}_f$ an approximate Riemann flux evaluated at face $f$
using left and right cell-average states. 

For face $f$ with outward unit normal $\nvec = (n_x, n_y)$, define the face-normal velocity $u_n = u\,n_x + v\,n_y$ and tangential velocity
$u_t = -u\,n_y + v\,n_x$. The Roe flux is
\begin{equation}
  \hat{\mathbf{F}}^{\mathrm{Roe}}_f
  = \tfrac{1}{2}\bigl[\mathbf{F}_n(\Q_L) + \mathbf{F}_n(\Q_R)\bigr]
  - \tfrac{1}{2}\sum_{k=1}^{3}
    |\tilde{\lambda}_k|\,\tilde{\alpha}_k\,\tilde{\mathbf{r}}_k,
\label{eq:roe}
\end{equation}
where $\mathbf{F}_n = \mathbf{F} n_x + \mathbf{G} n_y$ is the face-normal flux assembled from the perturbation-form fluxes \eqref{eq:swe_vars}, and the Roe averages are
\begin{equation}
  \tilde{u}_n
    = \frac{\sqrt{h_L}\,u_n^L + \sqrt{h_R}\,u_n^R}{\sqrt{h_L} + \sqrt{h_R}},
  \quad
  \tilde{u}_t
    = \frac{\sqrt{h_L}\,u_t^L + \sqrt{h_R}\,u_t^R}{\sqrt{h_L} + \sqrt{h_R}},
  \quad
  \tilde{c} = \sqrt{g\,\tfrac{h_L + h_R}{2}}.
\label{eq:roe_avgs}
\end{equation}
The eigenvalues $\tilde{\lambda}_{1,2,3} = \tilde{u}_n - \tilde{c},\, \tilde{u}_n,\, \tilde{u}_n + \tilde{c}$, wave strengths $\tilde{\alpha}_k$, and right eigenvectors $\tilde{\mathbf{r}}_k$ follow the standard 2D-SWE Roe decomposition \cite{toro2001shock}, with a Harten--Hyman entropy fix to prevent non-physical expansion shocks. Wall and inlet/outlet boundaries are handled via ghost cell states.

\subsection{Physics-Informed Neural Networks}
\label{sec:pinn}

PINNs \cite{raissi2019pinn} parameterise the solution as a neural network $\Q_\theta(\x, t)$, in our case an MLP that outputs the perturbation-form state $[\xi, uh, vh]$, and train by minimising: 
\begin{equation}
  \mathcal{L}_{\mathrm{PINN}} =
  \lambda_{\mathrm{PDEs}} \mathcal{L}_{\mathrm{PDEs}}
  + \lambda_{\mathrm{BC}} \mathcal{L}_{\mathrm{BC}}
  + \lambda_{\mathrm{IC}} \mathcal{L}_{\mathrm{IC}},
\label{eq:pinn_loss}
\end{equation}
where $\mathcal{L}_{\mathrm{PDEs}}$ penalises the pointwise strong-form PDE residual at collocation points. The BC and IC losses are $\mathcal{L}_{\mathrm{BC}}$ and $\mathcal{L}_{\mathrm{IC}}$, respectively. The weights $\lambda_{\mathrm{PDEs}}$, $\lambda_{\mathrm{BC}}$, and $\lambda_{\mathrm{IC}}$ balance the contributions of each loss term. While this is effective for smooth flows, it has critical deficiencies for hyperbolic systems such as the SWEs (see Table~\ref{tab:pinn_weaknesses}).

\begin{table}[htp]
\centering
\caption{Weaknesses of standard strong-form PINNs for the SWEs.}
\label{tab:pinn_weaknesses}
\begin{tabular}{lp{9cm}}
\toprule
\textbf{Weakness} & \textbf{Consequence for SWEs} \\
\midrule
Assumes smooth solutions
  & SWEs produce hydraulic jumps, bores, and wet-dry fronts where pointwise
    residuals are ill-defined \\
No conservation guarantee
  & Mass may not be conserved; physically inadmissible in hydrodynamics \\
No upwinding
  & Without entropy-satisfying fluxes, non-physical expansion shocks can emerge \\
Mesh agnostic
  & Cannot exploit boundary-conforming unstructured meshes \\
No wave-speed awareness
  & Hyperbolic information propagation at $c = \sqrt{gh} \pm u$ is not encoded \\
\bottomrule
\end{tabular}
\end{table}

\section{FVM-PINN Formulation}
\label{sec:method}

\subsection{Network Architecture}
\label{sec:network}

\begin{figure}[htp]
  \centering
  \includegraphics[width=0.9\textwidth]{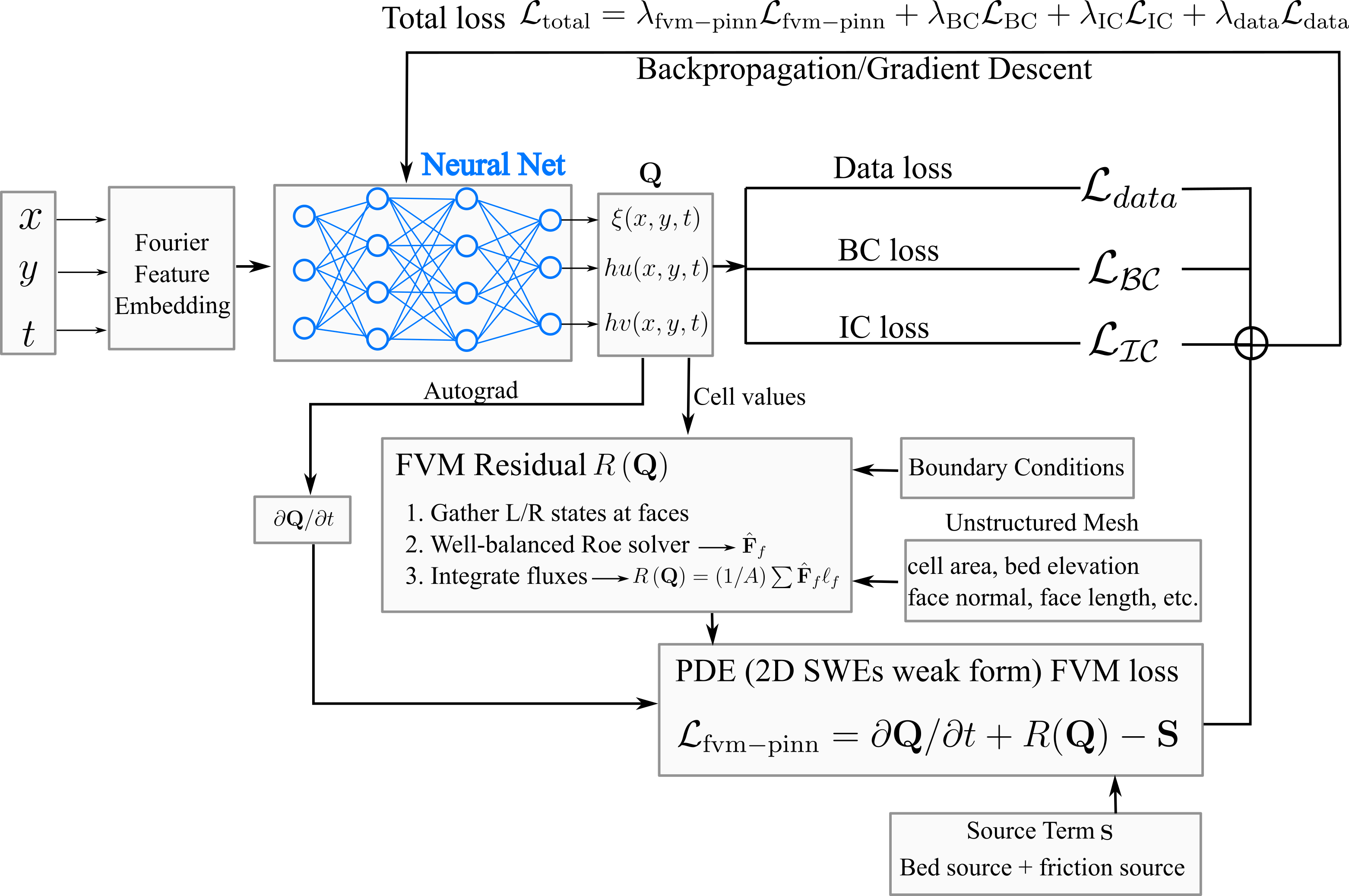}
  \caption{Schematic of the FVM-PINN architecture and loss components.}
  \label{fig:fvm_pinn_scheme}
\end{figure}

We parameterise the solution as a neural network with inputs $(x, y, t)$ and outputs $\Q_\theta = [\xi, uh, vh]$ in the perturbation form. The scheme of the FVM-PINN architecture is illustrated in Figure~\ref{fig:fvm_pinn_scheme}. Predicting $\xi$ rather than $h$ directly preserves the well-balanced property of the Roe solver: for the lake-at-rest condition, the network only has to match $\xi = 0$ and zero momentum, a quantitatively easier target than matching the spatially varying still-water depth $h_s(\x)$. To improve the learning of spatially varying features, we apply a Fourier feature embedding \cite{tancik2020fourier_features} before the first hidden layer: 
\begin{equation}
  \gamma(x, y, t) = \bigl[
    \cos(\mathbf{B}\,[x,y,t]^\top),\;
    \sin(\mathbf{B}\,[x,y,t]^\top)
  \bigr]^\top \in \R^{2m},
\label{eq:fourier}
\end{equation}
where $\mathbf{B} \in \R^{m \times 3}$ is fixed at initialisation with entries drawn from a zero-mean Gaussian distribution $\mathcal{N}(0, \sigma^2)$. $m$ is the number of Fourier features. The embedded vector feeds into an $L$-layer MLP with $\tanh$ activations and residual connections. A softplus activation on the recovered depth $h = \xi + h_s$ enforces $h \geq 0$ at output time, while inputs are $z$-score normalised on $(x, y, t)$.

\subsection{FVM-Informed Loss}
\label{sec:loss}

The total loss has four terms: 
\begin{equation}
  \mathcal{L}_{\mathrm{total}} =
  \lambda_{\mathrm{fvm-pinn}} \mathcal{L}_{\mathrm{fvm-pinn}}
  + \lambda_{\mathrm{BC}}  \mathcal{L}_{\mathrm{BC}}
  + \lambda_{\mathrm{IC}}  \mathcal{L}_{\mathrm{IC}}
  + \lambda_{\mathrm{data}} \mathcal{L}_{\mathrm{data}}.
\label{eq:total_loss}
\end{equation}
where $\mathcal{L}_{\mathrm{fvm-pinn}}$ is the FVM residual loss defined below. In contrast to the strong-form residual $\mathcal{L}_{\mathrm{PDEs}}$ in the standard PINN loss shown in equation~\eqref{eq:pinn_loss}, $\mathcal{L}_{\mathrm{fvm-pinn}}$ enforces the integral (weak) form of the SWEs through the FVM-PINN residual evaluated on cell-averaged states. $\mathcal{L}_{\mathrm{data}}$ is a data-fitting loss defined in Section~\ref{sec:data_guided} with weight $\lambda_{\mathrm{data}}$.

At each training step, we sample $n_t$ time levels $\{t_k\}$ uniformly from $[t_0, T]$, where $k = 0, \ldots, n_t-1$. For each time level, the network predicts cell-centre states $\Q_i = \Q_\theta(x_i, y_i, t_k)$ for all cells $i$. Based on the FVM discretisation and the Roe solver, the PDE loss for cell $i$ at time $t_k$ is
\begin{equation}
  R_i(t_k) =
  \frac{\partial \Q_i}{\partial t}\bigg|_{t_k}
  + \frac{1}{A_i}\sum_{f \in \partial \Omega_i}
    \hat{\mathbf{F}}^{\mathrm{Roe}}_f(\Q_{L(f)}, \Q_{R(f)}, \nvec_f)\,\ell_f
  - \mathbf{S}_i(t_k),
\label{eq:residual}
\end{equation}
where $\partial \Q_i / \partial t$ is computed via automatic differentiation through the network, and $\hat{\mathbf{F}}^{\mathrm{Roe}}$
is the Roe flux in equation~\eqref{eq:roe}. Assuming there are $n_c$ cells, the total FVM-PINN (PDE) loss is then the summation of the FVM loss for each cell and time level: 
\begin{equation}
  \mathcal{L}_{\mathrm{fvm-pinn}}
  = \frac{1}{n_t n_c}\sum_{k=1}^{n_t} \sum_{i=1}^{n_c}
    A_i \norm{R_i(t_k)}^2.
\label{eq:fvm_loss_full}
\end{equation}

The boundary conditions are enforced as follows. For wall boundaries, ghost-cell reflection enforces zero normal velocity. For open BCs (inlet discharge, exit water depth), the prescribed values replace the ghost state in the Roe flux computation. In addition to the enforcement of boundary conditions during the FVM discretisation, an additional soft BC penalty term $\mathcal{L}_{\mathrm{BC}}$ is applied at boundary face centres, which measures the deviation from the prescribed boundary values. Our experiments show that this additional penalty term improves convergence, especially at sparse data densities, by providing a stronger gradient signal at the boundaries. 

The initial condition loss $\mathcal{L}_{\mathrm{IC}}$ penalises the mismatch between the network prediction and the initial state $\Q_i^0$ at $t = t_0$:
\begin{equation}
  \mathcal{L}_{\mathrm{IC}}
  = \frac{1}{n_c}\sum_{i=1}^{n_c}
    \norm{\Q_\theta(x_i, y_i, t_0) - \Q_i^0}^2.
\label{eq:ic_loss}
\end{equation}

\subsection{Data-Guided Training}
\label{sec:data_guided}

The FVM-PINN, BC, and IC losses define a physics-only training objective. Mathematically, SWEs with proper boundary conditions and initial conditions define a well-posed problem. However, in Section~\ref{sec:data_ablation} we show that this physics-only formulation may converge to trivial solutions on realistic 2D problems due to rugged loss landscape. To overcome this, we introduce a data loss as an additional training signal to guide the optimizer out of shallow basins in the loss landscape. The data loss is defined as a weighted mean squared error between the network prediction and available observations or reference data:
\begin{equation}
  \mathcal{L}_{\mathrm{data}}
  = \frac{1}{N_d} \sum_{j=1}^{N_d}
    \norm{\Q_\theta(\x_j, t_j) - \Q_j^{\mathrm{obs}}}^2_{\mathbf{M}_j},
\label{eq:data_loss}
\end{equation}
where $\{(\x_j, t_j, \Q_j^{\mathrm{obs}})\}$ are observational or reference data points expressed in the same perturbation form as the
network output, and $\mathbf{M}_j$ is a per-point diagonal mask that selects which components are observed (e.g.\ velocity only, depth only, or both).

We consider two data sources that are commonly available in practice: 
\begin{enumerate}
  \item \textbf{Sparse measurements.} Point velocity or depth measurements from field sensors, ADCPs, or
    remote sensing. Typically $N_d \sim O(10^2)$ scattered points at one or a few time instants. The mask $\mathbf{M}_j$ reflects which variables the sensor records. 

  \item \textbf{Multi-snapshot reference data.} Cell-averaged states from a calibrated numerical model (e.g., SRH-2D) at several output time steps. This corresponds to the surrogate-from-simulator paradigm: the user has a trusted but expensive model and wants a fast neural surrogate. Typically $N_d = n_c \times n_{\mathrm{snap}}$ with $n_{\mathrm{snap}} \sim 5\text{--}10$.
\end{enumerate}

In addition to externally-provided data, a third practical option is an in-loop FVM ``teacher'': instead of loading reference data from files, the same well-balanced Roe FVM scheme used in $\mathcal{L}_{\mathrm{fvm-pinn}}$ is run as a forward solver from the IC at training time, producing $n_{\mathrm{snap}}$ snapshots that the network is distilled against. The marching forward in time is achieved by using a predictor-corrector (Heun) scheme. This is functionally a special case of multi-snapshot reference data - the snapshots happen to be self-generated rather than imported - and we show in Section~\ref{sec:data_ablation} (BIC-D vs BIC-H) that the two are interchangeable at the dense-data limit. The ``FVM teacher'' variant is most useful when an external solver run is expensive or unavailable.

The combined loss in equation~\eqref{eq:total_loss} with $\lambda_{\mathrm{data}} > 0$ serves a dual purpose: it pulls the optimizer out of the trivial low-momentum basin of the FVM-PINN loss (Section~\ref{sec:loss_landscape}) and anchors the network to observed physics. Crucially, the FVM-PINN loss remains active. It imposes conservation and upwinding at \emph{all} space-time points, not just the data locations, so the physics loss extrapolates conservation constraints to unobserved regions in a way pure data-fitting cannot. The relative weight of the two main losses, i.e., $\lambda_{\mathrm{fvm-pinn}}$ and $\lambda_{\mathrm{data}}$, is treated as a hyperparameter; the ablation in Section~\ref{sec:data_ablation} (BIC-G/H, comparing $\lambda_{\mathrm{fvm-pinn}} = 0$ to $\lambda_{\mathrm{fvm-pinn}} = 1$ at fixed $\lambda_{\mathrm{data}}$) shows that the FVM contribution is largest at sparse data densities and diminishes as data density grows.

Following standard PINN practice \cite{raissi2019pinn, lu2021deepxde}, we train in two phases: (1) Adam with optional step-decay learning rate for fast progress, and (2) L-BFGS for fine-grained convergence near the minimum.

\section{Scalability Strategies}
\label{sec:scalability}

Each FVM-PINN training step evaluates the residual in equation~\eqref{eq:residual} on every cell at every sampled time level, requiring $n_c \times n_t$ network forward passes per step together with automatic differentiation through them to obtain $\partial \Q / \partial t$. The autograd graph that records these operations for backpropagation occupies $O(n_c \times n_t)$ memory and becomes a bottleneck for larger meshes ($n_c > 10^4$) or longer simulations ($T \gtrsim 10^3$~s, in which many time levels $n_t$ must be sampled). Of the strategies we have implemented to address this, the one we demonstrate in this paper is sequential time-window decomposition; two additional capabilities, cell mini-batching with stencil expansion and gradient checkpointing, are available in the framework but are not included to reduce the paper's length.

\subsection{Sequential Time-Window Decomposition}
\label{sec:timewindow}

The time interval $[t_0, T]$ is partitioned into $N$ sub-intervals (``windows''). Each window trains an independent network on its own short horizon, and the initial condition for window $k$ is provided by the terminal state of window $k-1$:
\begin{equation}
  \Q^{(k)}_i(x_i, y_i, t_{k-1}) = \Q^{(k-1)}_\theta(x_i, y_i, t_{k-1}).
\label{eq:window_handoff}
\end{equation}
where $t_{k-1}$ is the boundary between window $k-1$ and window $k$ (the end of one, the start of the next). This handoff enforces temporal causality \cite{mattey2022sequential}. PINN training over a long single window tends to under-fit late-time states because the optimiser never gets a chance to focus on them; the partition converts a hard long-horizon problem into a chain of well-conditioned short-horizon problems. Warm-start initialisation (window $k$'s weights from window $k-1$) accelerates convergence. The cost is wall-clock because windows must be trained sequentially. The accuracy benefit is demonstrated in Section~\ref{sec:savannah} (SR-C/D).

\subsection{Other Available Options}
\label{sec:minibatch}

Two additional strategies are implemented in the codebase and were used in preliminary studies but are not the focus of this paper. \emph{Cell mini-batching with stencil expansion} samples a random subset $\mathcal{I}_{\mathrm{batch}}$ of cells at each step and includes the 1-ring stencil (all face-neighbours) in the forward pass; memory cost scales linearly in the sampling fraction $f$, and the stencil-expansion step is essential to preserve the exact cell-flux balance the Roe solver requires. \emph{Gradient checkpointing} \cite{chen2016gradient_checkpoint} recomputes intermediate activations during the backward pass, reducing the peak memory of the per-step autograd graph from $O(L n_c)$ to $O(\sqrt{L n_c})$ at the cost of one additional forward pass per gradient step, where $L$ is the depth of the MLP. Both are orthogonal to time-windowing and to each other. 

\section{Numerical Experiments}
\label{sec:experiments}

We organise the experiments in four parts: (1)~validation on problems with exact or well-characterised reference solutions
    (\S\ref{sec:dambreak}--\ref{sec:bump}); (2)~data-guided training ablation on a 2D benchmark (\S\ref{sec:data_ablation});
(3)~real-world application to the Savannah River (\S\ref{sec:savannah}); (4)~scalability strategy comparison (\S\ref{sec:strategy_comparison}).

All experiments use double precision (float64) and the two-phase Adam + L-BFGS training protocol. Default network and training hyperparameters are listed in Table~\ref{tab:hyperparams} (\ref{app:hyperparams}). All implementations used PyTorch. Random seeds for PyTorch, NumPy, and CUDA are fixed across all runs to support reproducibility; the specific seed values are documented in the per-case YAML configurations released with the companion code repository. The wall time, obtained on a computer with NVIDIA Quadro RTX 4000 GPU, is reported to show the computational cost.

\subsection{Case 1: 1D Dam-Break Validation}
\label{sec:dambreak}

We validate the FVM-PINN formulation on a 1D dam-break problem with an exact analytical solution \cite{stoker1957water, toro2001shock}. The domain is $[0, 20]$~m with the dam at $x = 10$~m. Initial depths are $h_L = 2.0$~m and $h_R = 0.5$~m with zero initial velocities. The simulation end time is $T = 1$~s. The domain is discretised into 100 equal-width cells (modelled as a thin 2D strip of width 0.2~m to exercise the 2D FVM machinery). No measurement or reference data is used. The training has $5000$ Adam steps with $\eta = 10^{-3}$ decayed by $\times 0.9$ every $1000$ steps, $5$ random time samples per step (no L-BFGS, since the 1D residual is already small).

\begin{figure}[h!]
  \centering
  \includegraphics[width=0.9\textwidth]{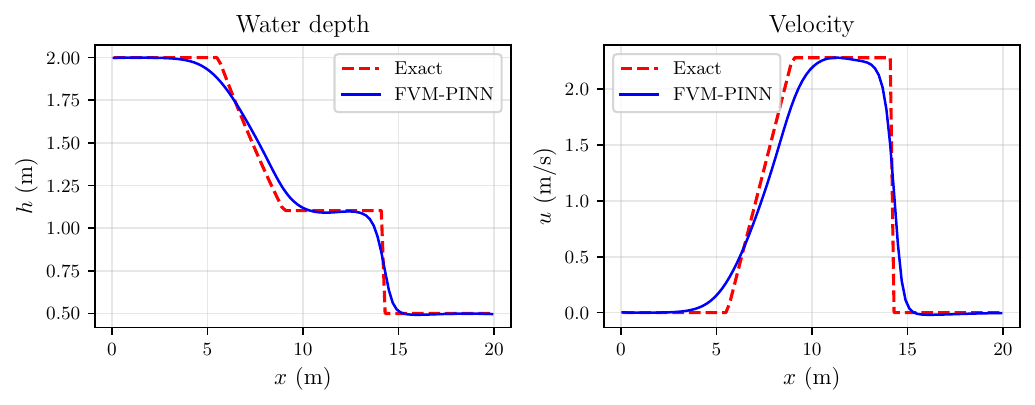}
  \caption{Case 1: 1D dam-break. FVM-PINN predictions (blue) vs exact solution (red dashed) at $t=1$~s. (a) water depth $h(x)$, (b) velocity $u(x)$.}
  \label{fig:dambreak}
\end{figure}

Figure~\ref{fig:dambreak} shows the predicted and exact water depth and velocity profiles at $t = 1$~s. The network correctly captures the left rarefaction fan, the contact wave, and the right-moving shock. Table~\ref{tab:dambreak} quantifies the errors in $L_2$ and $L_\infty$ norms, showing good agreement with the exact solution. For this simple case, the FVM-PINN formulation demonstrates excellent performance in capturing the essential features of the flow dynamics.

\begin{table}[htp]
\centering
\caption{Case 1: 1D dam-break accuracy at $t = 1$~s ($n_c = 100$ cells).}
\label{tab:dambreak}
\begin{tabular}{lcccc}
\toprule
 & $L_2(h) (m)$ & $L_2(u) (m/s)$ & $L_\infty(h) (m)$\\
\midrule
FVM-PINN & 5.91e-2 & 1.96e-1 & 2.46e-1 \\
\bottomrule
\end{tabular}
\end{table}

\subsection{Case 2: Transcritical Flow Over a Bump}
\label{sec:bump}

In this case, a $25$~m channel has a parabolic bed bump centred at $x = 10$~m. Subcritical upstream flow transitions to supercritical over the bump crest, then jumps back to subcritical downstream. This case tests both the bed-slope source-term split in equation \eqref{eq:source_split} (well-balanced reconstruction) and Manning friction. Specific inlet discharge $q = 0.18$~m$^2$/s, exit water-surface elevation $w_s = 0.33$~m, Manning's $n = 0.02$. Training is over $T = 72$~s. For FVM-PINN with data guide, the inner FVM solver (teacher) generates a $51$-snapshot trajectory which the network is distilled against. For FVM-PINN without data guide, the network is trained purely on the PDEs, BC, and IC losses. The reference solution is the analytical steady state of \citeA{goutal1997monitoring}. The training has $3000$ Adam steps with $\eta = 10^{-3}$ decayed by $\times 0.9$ every $2000$ steps, followed by $200$ L-BFGS steps. Other hyperparameters are shown in Table~\ref{tab:hyperparams}.

\begin{figure}[htp]
  \centering
  \includegraphics[width=0.9\textwidth]{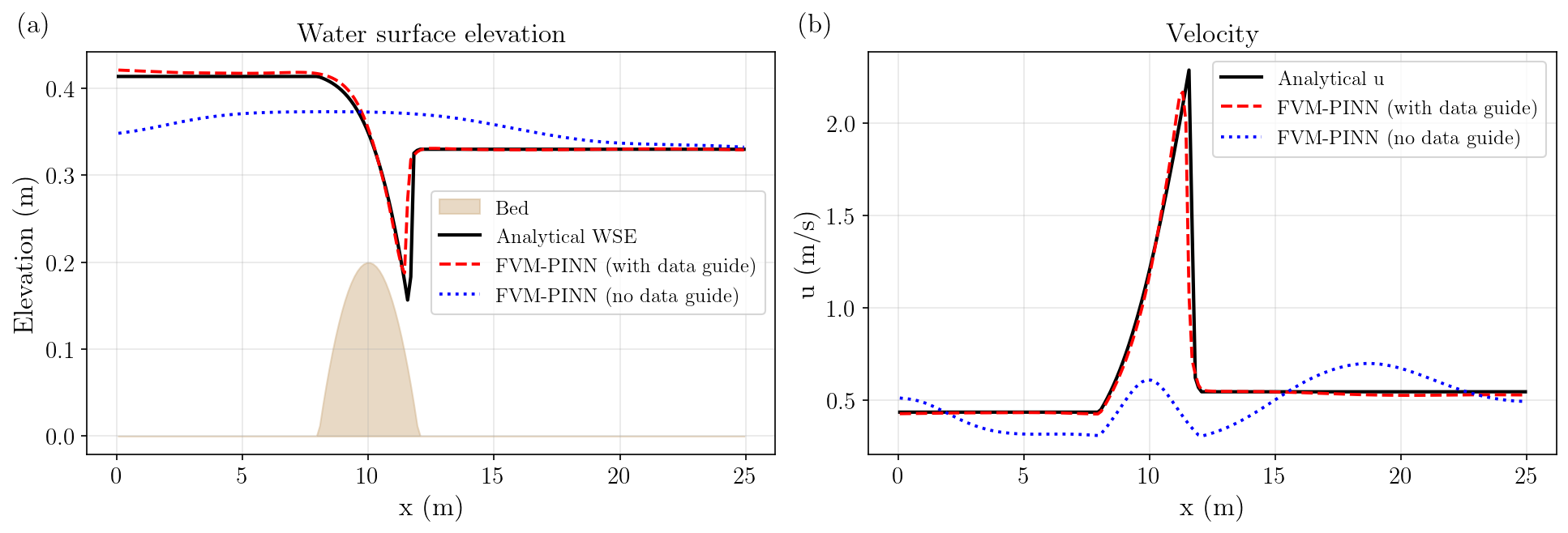}
  \caption{Case 2: WSE profile over a parabolic bed bump at the final time step (steady state). (a) WSE, (b) velocity.}
  \label{fig:bump}
\end{figure}

Figure~\ref{fig:bump} shows the steady water surface elevation (WSE) profile at the final time step. FVM-PINN with data guide correctly captures the subcritical-to-supercritical transition over the crest and the hydraulic jump downstream, achieving $L_2(h) = 1.31{\times}10^{-2}$ m and $L_2(u) = 1.03{\times}10^{-1}$ m/s against the analytical solution. Without data guide, the network fails to converge to the correct solution and instead collapses to a trivial state where the WSE and velocity are almost uniform. This highlights the importance of data guidance for training stability and convergence in realistic 2D problems, which we systematically study in the next section.

\subsection{Case 3: Data-Guided Training Ablation}
\label{sec:data_ablation}

This is the central experiment of the paper. We use a 2D block-in-channel (BIC) benchmark (15~m $\times$ 5~m rectangular channel with a block obstruction, 1326 unstructured cells) to systematically study the role of data in FVM-PINN training.

The setup of the case is as follows. The block is 1~m wide and 0.5~m long, centred at $(x, y) = (5, 2.5)$~m. The initial condition is a quiescent state with uniform depth $h = 0.4$~m. Inlet discharge $Q = 0.38$~m$^3$/s, exit depth $h = 0.4$~m, Manning's $n = 0.03$, simulation time $T = 360$~s. The channel bed is flat with an elevation $z_b(x) = 0$ m. The block creates a truly 2D flow pattern with separation, recirculation, and wake effects. The SRH-2D steady-state solution at $t = 360$~s serves as the ground
truth. All runs except BIC-D use the same network (6 layers $\times$ 128 units, 64 Fourier features) and the same standard-trainer schedule (8000 Adam + 500 L-BFGS steps); BIC-D uses the FVM teacher variant introduced in Section~\ref{sec:method}, which adds an inner FVM trajectory data generator. All metrics in Table~\ref{tab:BIC_ablation} are computed against SRH-2D.

\begin{figure}[htp]
  \centering
  \includegraphics[width=\textwidth]{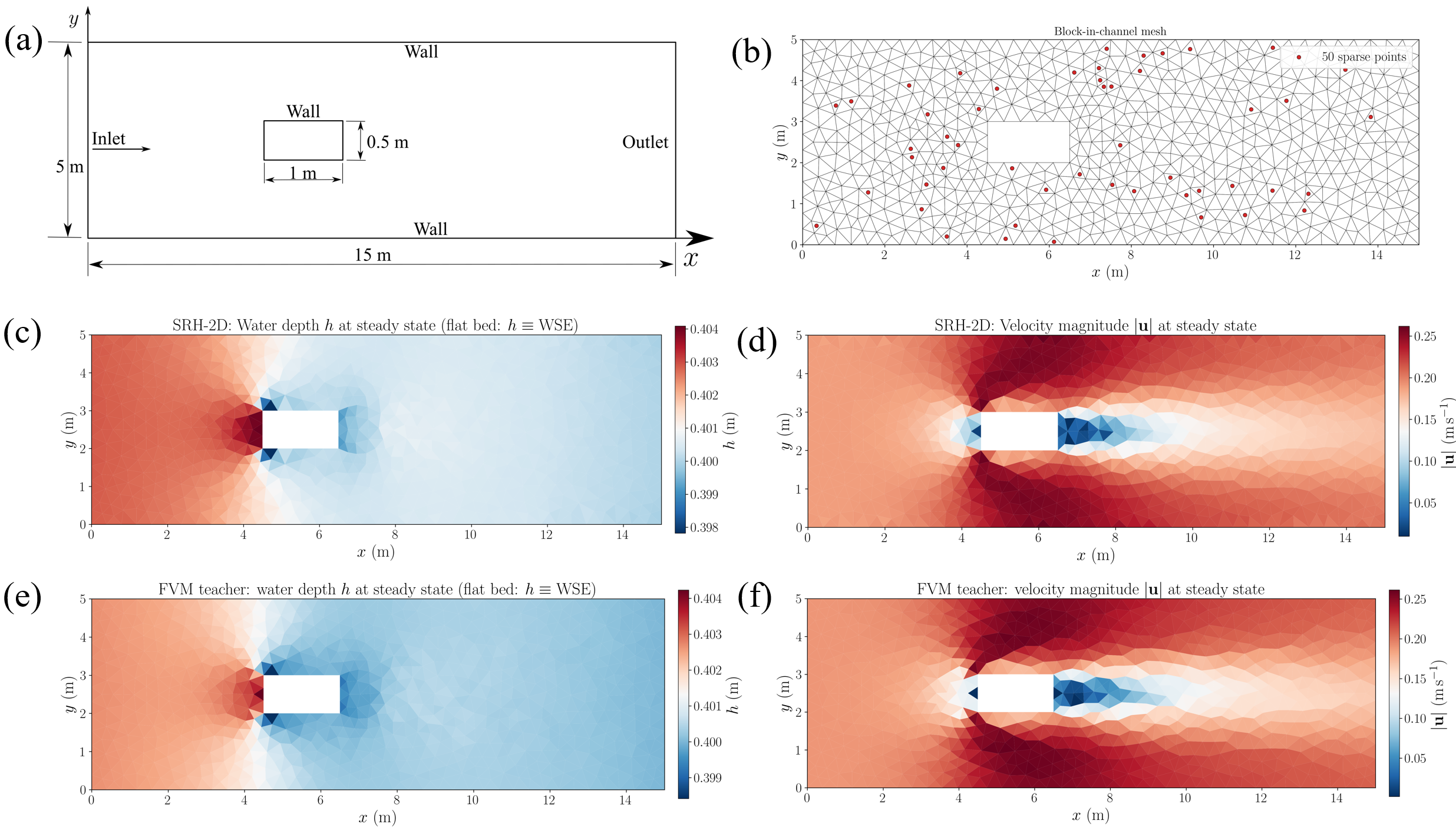}
  \caption{Case 3: Block-in-channel benchmark. (a) Case setup, (b) Unstructured mesh and 50 sparse data points, (c) Water depth result from SRH-2D, (d) Velocity result from SRH-2D, (e) Water depth result from FVM teacher, (f) Velocity result from FVM teacher.}
  \label{fig:block_in_channel}
\end{figure}

Table~\ref{tab:BIC_ablation} summarises eight ablation run configurations. The matrix sweeps three axes: data quantity (none, 50 sparse, 200 sparse, dense SRH-2D/teacher snapshots, and a combined sparse + dense block), data quality (clean vs.\ 5\% Gaussian noise on the sparse rows), and the role of the FVM-PINN residual itself (the data-only baselines BIC-G and BIC-H, with $\lambda_{\mathrm{fvm-pinn}} = 0$).

\begin{table}[htp]
\centering
\caption{Case 3: Data-guidance ablation on block-in-channel. Errors are computed against SRH-2D at $t$ = $360$~s. All runs use the same network and the same Adam + L-BFGS schedule except BIC-D, which is guided by FVM teacher-generated solution trajectory snapshots (Section~\ref{sec:method}); BIC-G and BIC-H disable the FVM-PINN residual ($\lambda_{\mathrm{fvm-pinn}}$ = 0) to isolate the contribution of the data loss.}
\label{tab:BIC_ablation}
\begin{tabular}{lp{4.4cm}lccr}
\toprule
\textbf{Run} & \textbf{Data mode} & \textbf{$N_d$}
  & $L_2(h)$ (m) & $L_2(|\mathbf{u}|)$ (m/s) & \textbf{Wall time (s)} \\
\midrule
BIC-A & None (physics only)            & 0             & 2.48e-3 & 1.66e-1 & 6,999  \\
BIC-B & Sparse velocity (200 pts)      & 200           & 4.51e-3 & 7.67e-3 & 6,909  \\
BIC-C & Sparse velocity (50 pts)       & 50            & 4.15e-3 & 2.24e-2 & 6,228  \\
BIC-D & Teacher-generated trajectory   & 31 snapshots  & 4.21e-4 & 6.88e-3 & 18,970 \\
BIC-E & Sparse velocity (200 pts) + dense snapshots       & $200 + 5 n_c$ & 3.25e-4 & 7.00e-3 & 6,051  \\
BIC-F & Sparse velocity (200 pts) + 5\% noise             & 200           & 4.99e-3 & 1.18e-2 & 6,043  \\
BIC-G & Sparse velocity (200 pts), $\lambda_{\mathrm{fvm-pinn}}{=}0$ & 200      & 1.53e-3 & 9.95e-3 & 5,965  \\
BIC-H & Dense velocity, $\lambda_{\mathrm{fvm-pinn}}{=}0$  & $5 n_c$  & 2.74e-4 & 6.32e-3 & 5,778  \\
\bottomrule
\end{tabular}
\end{table}

\begin{figure}[htp]
  \centering
  \includegraphics[width=\textwidth]{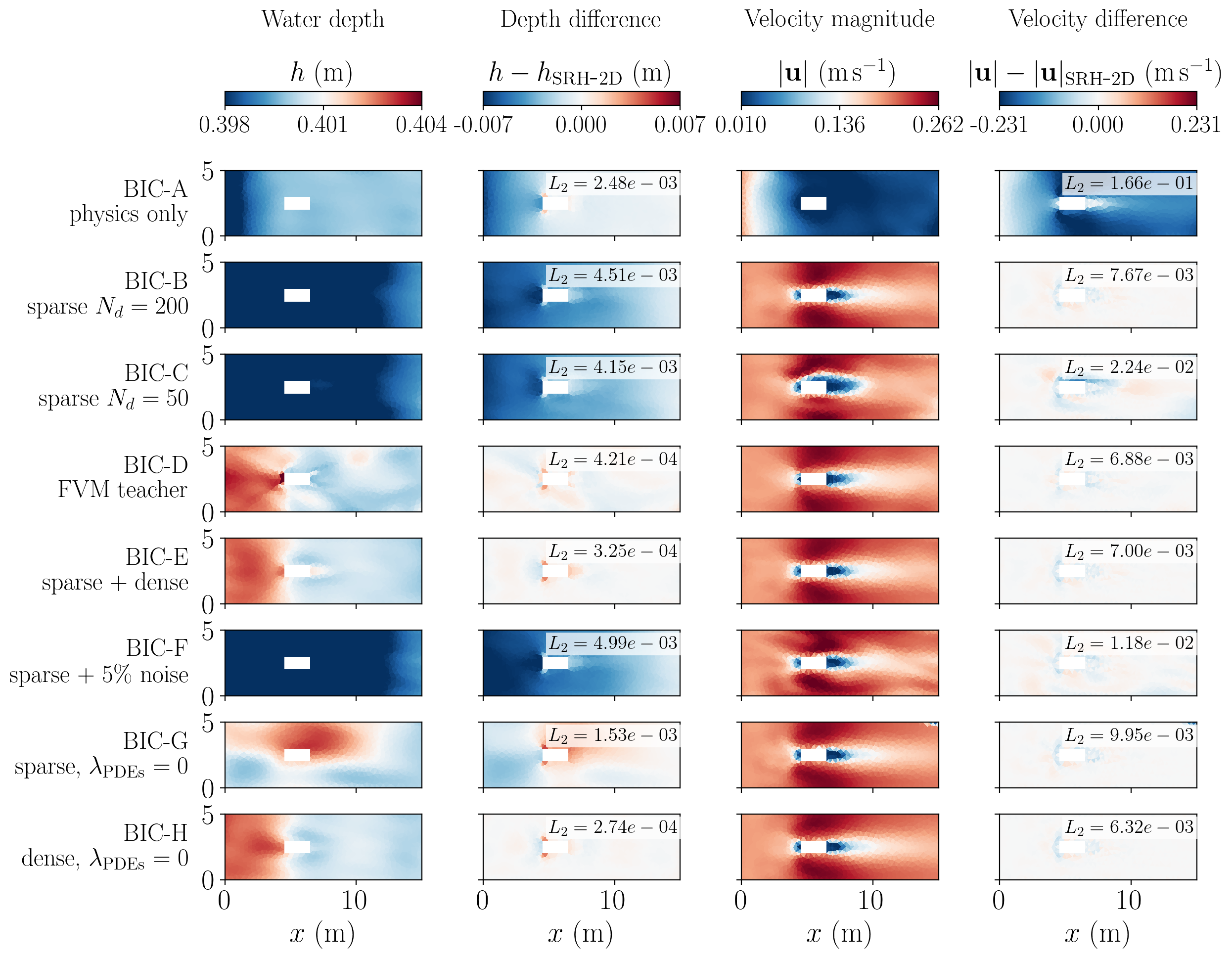}
  \caption{Case 3: Water depth and velocity contours at steady state for the eight ablation runs. The reference solution is from SRH-2D. The four columns correspond to water depth, difference in water depth from SRH-2D, velocity magnitude, and difference in velocity magnitude from SRH-2D. The rows correspond to the eight runs in Table~\ref{tab:BIC_ablation}.}
  \label{fig:BIC-h-velo-contours}
\end{figure}

\begin{figure}[htp]
  \centering
  \includegraphics[width=0.9\textwidth]{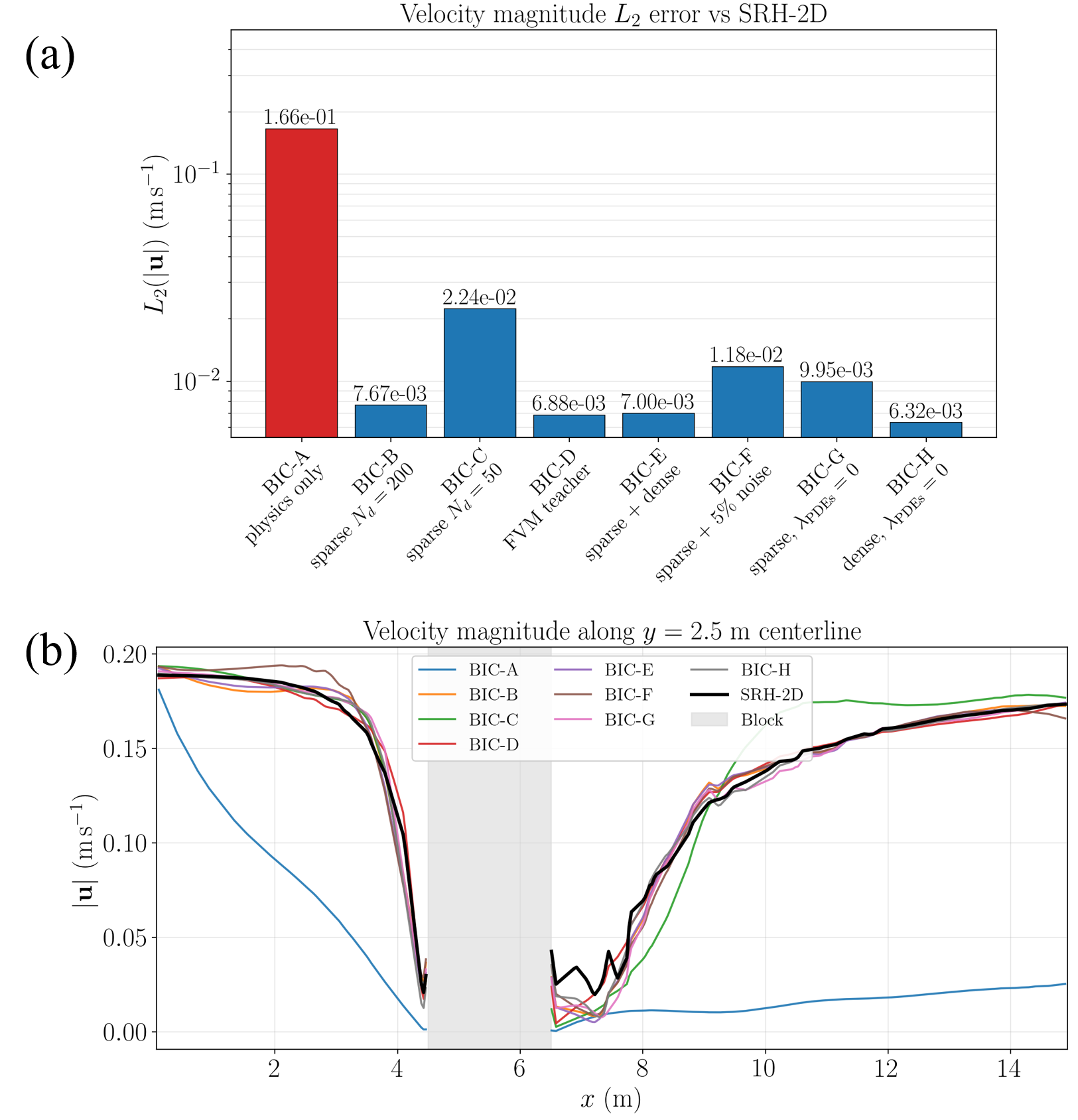}
  \caption{Case 3: Ablation summary and comparison: (a) $L_2(|\mathbf{u}|)$ for the eight runs in Table~\ref{tab:BIC_ablation} and (b) WSE profiles along the channel centreline for the same runs. The reference solution is from SRH-2D.}
  \label{fig:BIC_ablation_bar_profiles}
\end{figure}

By analysing the results in Table~\ref{tab:BIC_ablation}, the corresponding flow fields shown in Fig.~\ref{fig:BIC-h-velo-contours}, and the bar chart and WSE profiles in Fig.~\ref{fig:BIC_ablation_bar_profiles}, we draw the following conclusions:
\begin{itemize}
  \item Physics-only training fails (BIC-A). Without data guide, the optimizer converges to a near-uniform depth ($h \in [0.396, 0.400]$~m and near-zero momentum (with the only exception of the inlet region). The FVM-PINN loss is in fact small at this state. The trivial zero-momentum field nearly satisfies cell-wise conservation. But $L_2(|\mathbf{u}|) = 0.166$ confirms it bears no resemblance to the true flow field. Section~\ref{sec:loss_landscape} provides a quantitative loss-landscape diagnostic showing this is a shallow basin of the FVM-PINN loss that only the data loss can break out of.

  \item Sparse measurements rescue the solution (BIC-B). Just 200 random velocity measurements drop $L_2(|\mathbf{u}|)$ from $0.166$ to $7.67{\times}10^{-3}$, a $22\times$ reduction on the metric that matters most. The FVM-PINN  (PDEs) loss then enforces conservation and upwinding at all space-time points, propagating the data signal across the entire domain.

  \item Graceful degradation with sparsity (BIC-C). Reducing the measurement count from 200 to 50 ($4\times$ sparser) only triples the velocity error (from $7.67{\times}10^{-3}$ to $2.24{\times}10^{-2}$) and the qualitative wake structure is still recovered. Useful data densities for practice, e.g., ADCP transects with $O(10^2)$ measurements, are well within this regime.

  \item FVM teacher distillation matches direct supervision (BIC-D). Distilling the network against an internally-generated FVM trajectory of $31$ snapshots achieves $L_2(h) = 4.21{\times}10^{-4}$ and $L_2(|\mathbf{u}|) = 6.88{\times}10^{-3}$. This is essentially indistinguishable from BIC-H (direct supervision against the SRH-2D snapshots; see below), confirming that an in-loop FVM teacher and a pre-computed SRH-2D anchor are interchangeable as supervision sources at the dense-data limit. BIC-D is about $3{\times}$ slower because each training step generates a fresh teacher snapshot.

  \item Combined sparse + dense data gives state-of-the-art $h$ accuracy (BIC-E). Adding both 200 sparse velocity measurements at the steady state and dense SRH-2D snapshots over the transient brings $L_2(h)$ to $3.25{\times}10^{-4}$, comparable to the dense-only data-fitting baseline BIC-H ($2.74{\times}10^{-4}$). The sparse and dense supervision are complementary in the sense that sparse pins the final state while dense constrains the path to it.

  \item Noise sensitivity is real but bounded (BIC-F). Adding $5\%$ Gaussian noise (in fraction of $\max|\mathbf{u}|$) to the same 200 sparse measurements degrades $L_2(|\mathbf{u}|)$ from $7.67{\times}10^{-3}$ (BIC-B) to $1.18{\times}10^{-2}$, a $\sim$50\% increase. This is well above the BIC-G data-only baseline of $9.95{\times}10^{-3}$ but still $\sim$14$\times$ below BIC-A. Noise erodes some of the data-guidance benefit but does not push the optimizer back into the trivial basin.

  \item The FVM-PINN loss is most valuable in the sparse regime. Comparing the no-physics ablations to their physics-on counterparts decomposes the contribution of $\lambda_{\mathrm{fvm-pinn}}\mathcal{L}_{\mathrm{fvm-pinn}}$. At sparse data, the FVM-PINN loss reduces $L_2(|\mathbf{u}|)$ by $\sim$23\% ($\mathrm{BIC{-}G} = 9.95{\times}10^{-3} \to \mathrm{BIC{-}B} = 7.67{\times}10^{-3}$). At dense data, BIC-H ($\lambda_{\mathrm{fvm-pinn}} = 0$, dense data only) achieves $L_2(|\mathbf{u}|) = 6.32{\times}10^{-3}$ and $L_2(h) = 2.74{\times}10^{-4}$, marginally better than the physics-on combined-data run BIC-E ($7.00{\times}10^{-3}$ and $3.25{\times}10^{-4}$). With dense supervision on every cell, the data loss already provides the gradient signal the FVM-PINN loss would supply, so the loss's marginal contribution effectively vanishes. The implication is therefore not ``the FVM-PINN loss is essential'' but ``the FVM-PINN loss is essential at low data density and a diminishing-returns regulariser at high data density''. In practice, sparse field measurements are the common case, so the FVM-PINN loss is most useful where it is most needed.
\end{itemize}

\subsection{Case 4: Savannah River Application}
\label{sec:savannah}

\begin{figure}[htp]
  \centering
  \includegraphics[width=0.95\textwidth]{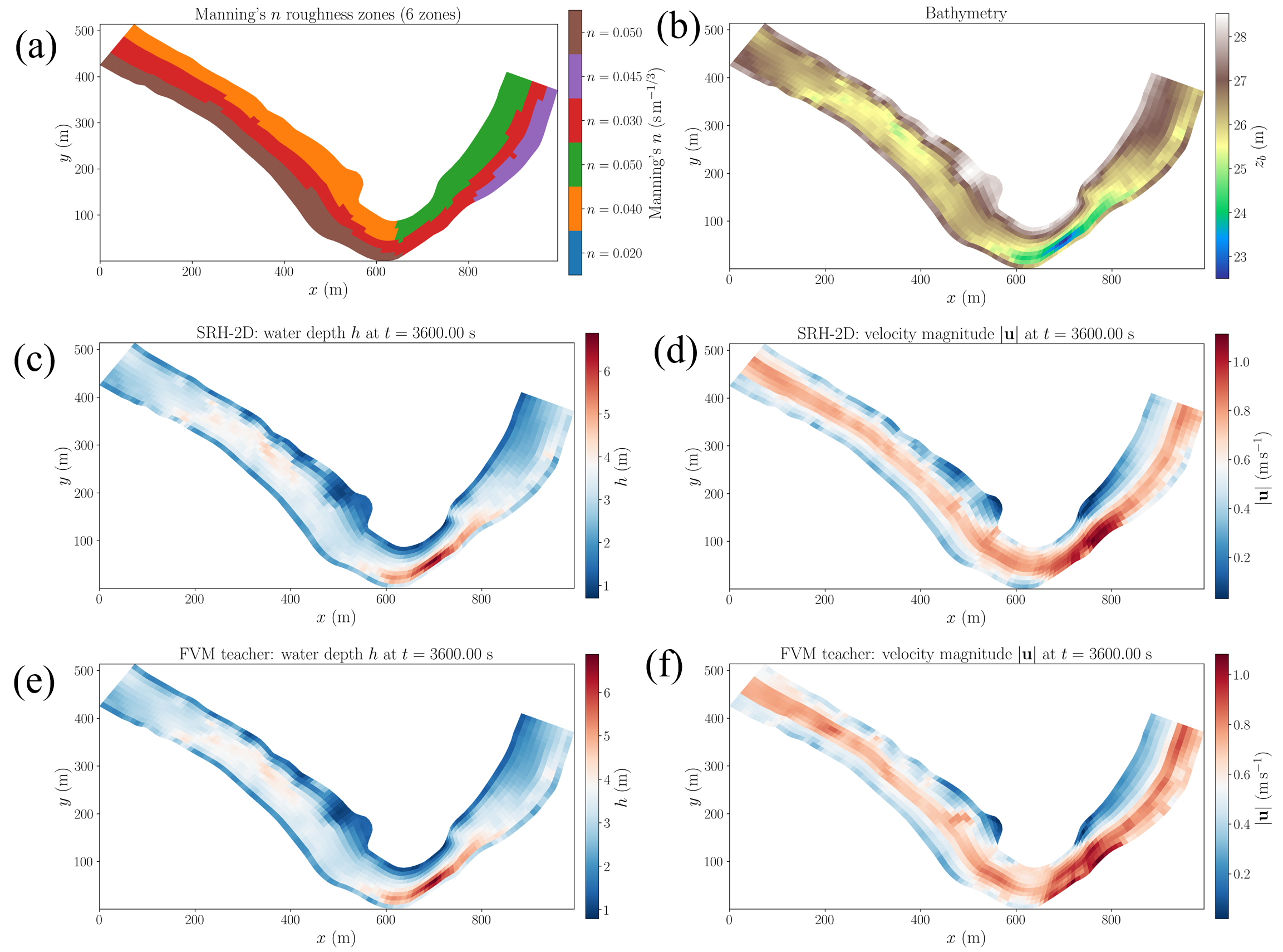}
  \caption{Case 4: Savannah River. (a) Manning's roughness zones, (b) Bathymetry, (c) Water depth from SRH-2D at steady state, (d) Velocity magnitude from SRH-2D at steady state, (e) Water depth from FVM teacher, and (f) Velocity magnitude from FVM teacher.}
  \label{fig:savannah_setup}
\end{figure}

We apply the full framework to a $\sim$1~km reach of the Savannah River, discretised into 1,306 unstructured cells with non-trivial bathymetry (bed elevation 22.5--28.5~m) and five Manning roughness zones ($n = 0.02$--$0.05$). Figure~\ref{fig:savannah_setup} shows the domain setup, bathymetry, and steady state flow field results from SRH-2D and FVM teacher. Boundary conditions are as follows: inlet discharge $Q = 187.4$~m$^3$/s distributed across the inlet by conveyance, exit water surface elevation $w_s = 29.3$~m. The training interval is $[0, 3600]$~s and the flow reaches steady state at the end. The initial condition is a flat water surface at $w_s = 29.3$~m, a deliberately uninformed cold start that the network has to recover from. SRH-2D provides reference snapshots at five times $t \in \{720, 1440, 2160, 2880, 3600\}$~s, all used as anchor data ($\xi$, $uh$, $vh$ on every cell). This case represents a practical surrogate-construction workflow: a modeler has a calibrated SRH-2D model and wants a fast neural surrogate for rapid scenario exploration, sensitivity analysis, or real-time forecasting. The data of this case was originally from \citeA{LeeEtAl2018} and has been used in surrogate modelling studies such as \citeA{liu2024bathymetry_inversion} and \citeA{liu2025uswe}.

Table~\ref{tab:savannah} reports seven run configurations. SR-A is the physics-only baseline (no SRH-2D anchor, no sparse measurements, no teacher trajectory - a real-world counterpart to BIC-A). SR-B is the single-network standard trainer with the dense SRH-2D anchor; SR-C and SR-D apply the time-window strategy of Section~\ref{sec:timewindow} with $5$ and $10$ sub-networks, respectively. SR-E is the teacher-distillation variant introduced in Section~\ref{sec:method}: instead of consuming externally-provided SRH-2D anchor data, an in-loop FVM solver generates a reference trajectory from the same flat-WSE cold start, and the network is distilled against the cached snapshots (analogous to BIC-D in Case 3). SR-F uses only $200$ sparse velocity measurements at $t = 3600$~s (no dense anchor); SR-G is a data-only ablation ($\lambda_{\mathrm{fvm-pinn}} = 0$) using the same anchor data as SR-B. All errors are computed against SRH-2D results at $t = 3600$~s.

\begin{table}[htp]
\centering
\caption{Case 4: Savannah River results at $t$ = $3600$~s. Errors are $L_2$ against SRH-2D results. $n_c$ denotes the number of mesh cells (here $n_c$ = $1{,}306$); $N_d$ counts cell-wise data values across all anchor / teacher snapshots, except SR-F which uses point-wise sparse velocity data. SR-A is the physics-only baseline (no data of any kind), included to mirror the BIC-A vs.\ BIC-B--H contrast in Case 3.}
\label{tab:savannah}
\setlength{\tabcolsep}{4pt}
\resizebox{\textwidth}{!}{%
\begin{tabular}{llp{4.4cm}lccr}
\toprule
\textbf{Run} & \textbf{Strategy} & \textbf{Data mode} & \textbf{$N_d$}
  & $L_2(h)$ (m) & $L_2(|\mathbf{u}|)$ (m/s) & \textbf{Wall time (s)} \\
\midrule
SR-A & Standard    & None (physics only)                                & 0               & 1.44e-1     & 6.00e-1     & 7,381  \\
SR-B & Standard    & SRH-2D anchor (5 snap.)                            & $5 n_c$         & 7.48e-2 & 3.20e-2 & 8,020  \\
SR-C & Window(5)   & SRH-2D anchor (5 snap.)                            & $5 n_c$         & 6.77e-2 & 2.04e-2 & 17,200 \\
SR-D & Window(10)  & SRH-2D anchor (5 snap.)                            & $5 n_c$         & \textbf{4.49e-2} & \textbf{1.31e-2} & 23,381 \\
SR-E & Teacher     & FVM trajectory (31 snap.) + SRH-2D anchor (5 snap.) & $(31{+}5)\,n_c$ & 7.49e-2 & 3.56e-2 & 14,263 \\
SR-F & Standard    & Sparse velocity (200 pts) at $t = T$               & 200             & 1.83e-1 & 7.23e-2 & 7,065  \\
SR-G & Standard, $\lambda_{\mathrm{fvm-pinn}}{=}0$ & SRH-2D anchor (5 snap.) & $5 n_c$     & 7.33e-2 & 3.16e-2 & 7,897  \\
\bottomrule
\end{tabular}%
}
\end{table}

\begin{figure}[htp]
  \centering
  \includegraphics[width=0.95\textwidth]{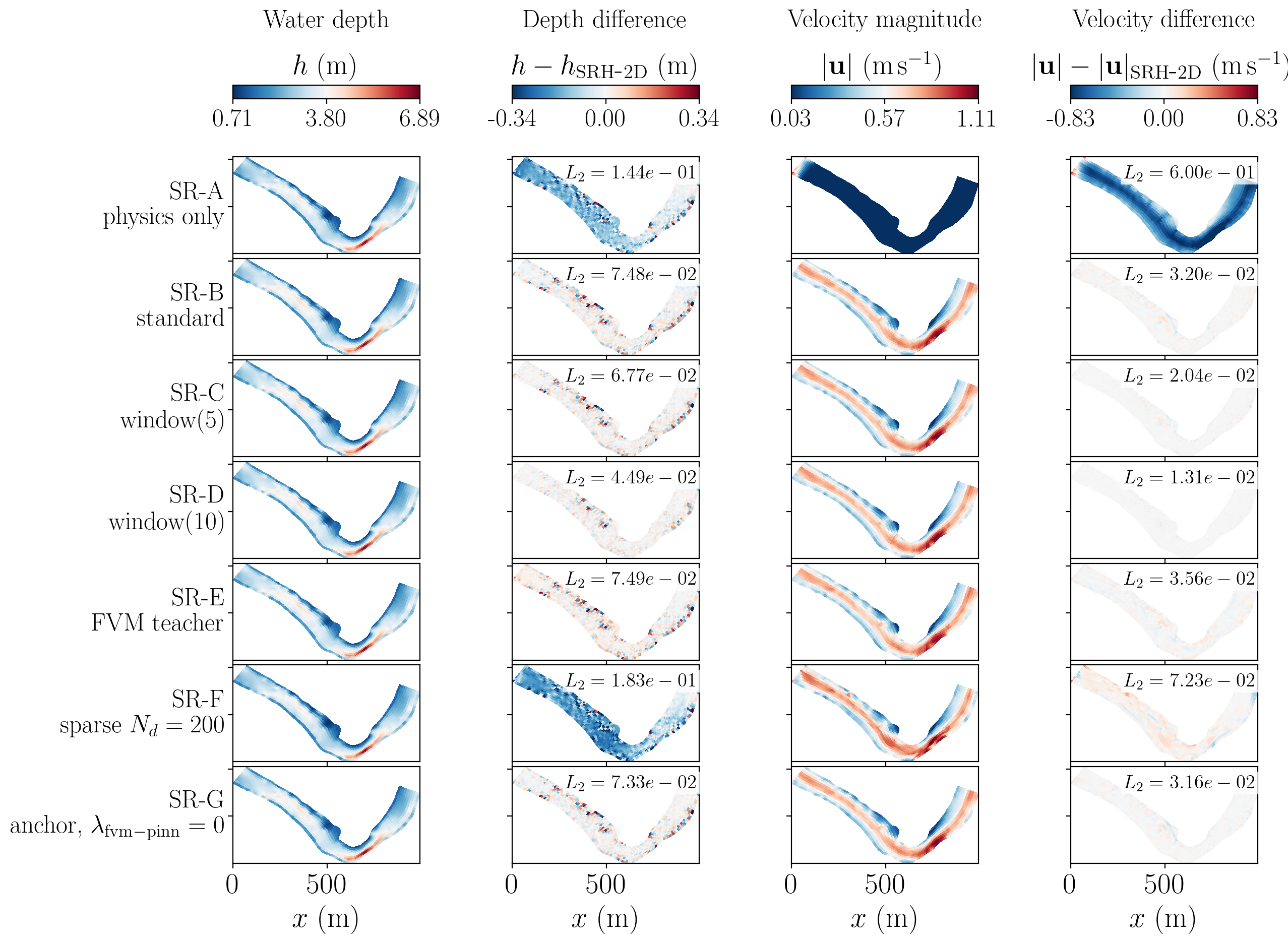}
  \caption{Case 4: Water depth and velocity contours at steady state for the seven ablation runs. The reference solution is from SRH-2D. The four columns correspond to water depth, difference in water depth from SRH-2D, velocity magnitude, and difference in velocity magnitude from SRH-2D. The rows correspond to the runs in Table~\ref{tab:savannah}.}
  \label{fig:savannah}
\end{figure}

\begin{figure}[htp]
  \centering
  \includegraphics[width=0.8\textwidth]{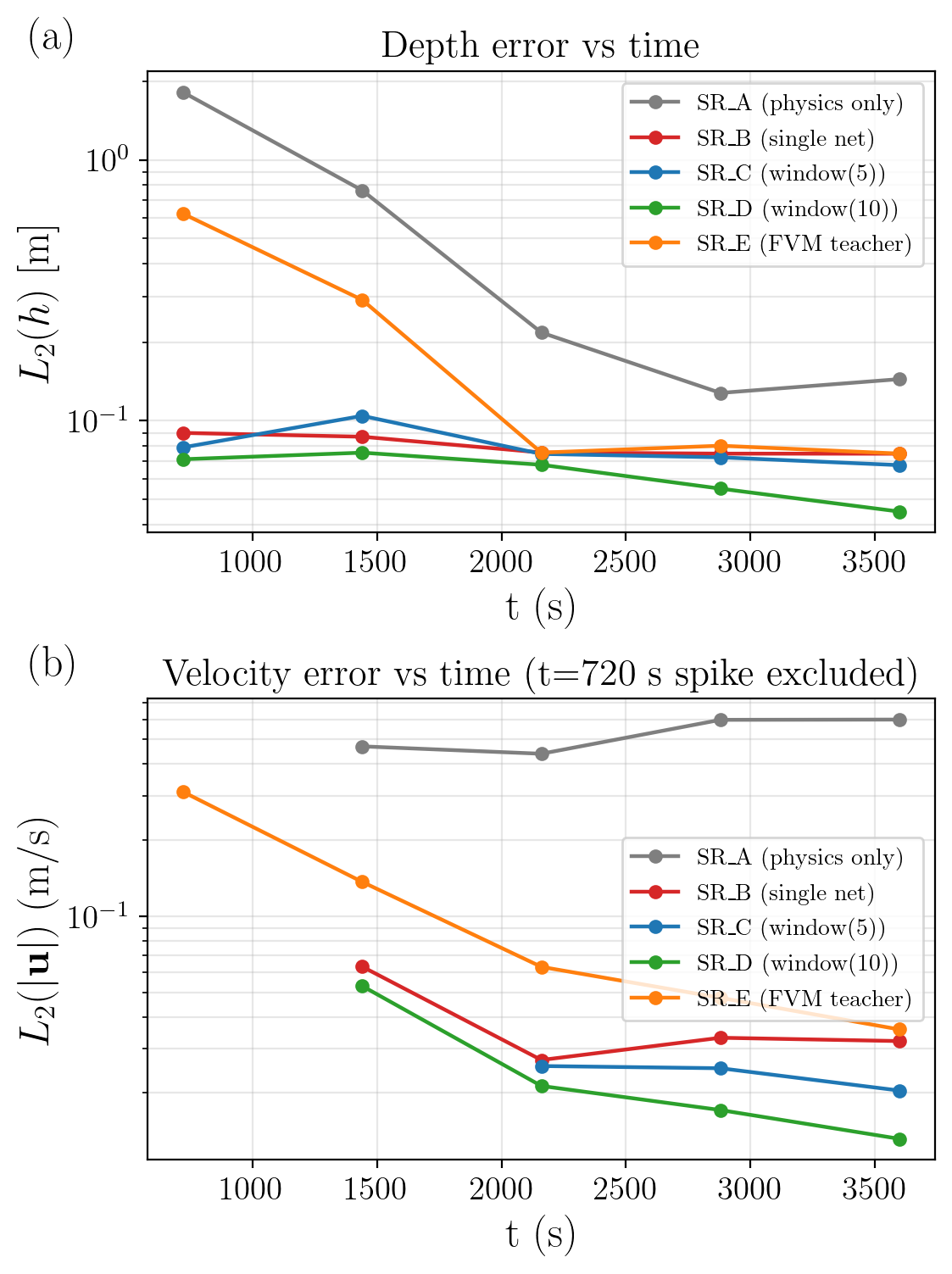}
  \caption{Case 4: $L_2$ error at each reference time ($t \in \{720, 1440, 2160, 2880, 3600\}$~s) for the physics-only baseline (SR-A), the standard trainer (SR-B), Window(5) (SR-C), Window(10) (SR-D), and the FVM teacher (SR-E). The windowed trainers' improvement over the single-network baseline grows monotonically with $t$ thanks to progressive IC handoff, while the single-network trainer's error is nearly constant in $t$.}
  \label{fig:savannah_l2}
\end{figure}

The Savannah River results confirm that FVM-PINN can produce an accurate neural surrogate for a real-world river reach. The results are shown in Table~\ref{tab:savannah} and Figure~\ref{fig:savannah_l2}. We have the following findings:
\begin{itemize}
  \item Physics-only training fails on the real reach. SR-A, which uses no data guidance whatsoever ($\lambda_{\mathrm{data}} = 0$ with the SRH-2D anchor disabled), is included as a real-world counterpart to BIC-A. As in Case 3, the absence of any data signal lets the network drift toward a near-trivial state during the long $3600$~s transient: the FVM residual, the inlet/exit BC loss, and the flat-WSE IC alone do not provide enough information to pin the temporal trajectory. This run is reported here to motivate the data-guided variants SR-B--SR-G; the rest of this discussion assumes some form of data guidance is available.

  \item Time windowing helps monotonically with the number of windows. At $t = 3600$~s, $L_2(h)$ drops from $0.075$ (SR-B, single network) to $0.068$ (SR-C, $5$ windows) to $0.045$ (SR-D, $10$ windows), a $\sim$$1.7\times$ accumulated improvement; $L_2(|\mathbf{u}|)$ drops by $\sim$$2.4\times$ over the same sequence. The improvement increases over time: the windowed trainers are within a few percent of the standard trainer at $t = 720$~s but open up a $2\times$ gap by $t = 3600$~s (Fig.~\ref{fig:savannah_l2}). The mechanism is the IC handoff (equation~\eqref{eq:window_handoff}): each window starts from a state already fitted by its predecessor, converting a hard 3600-s problem into a chain of well-conditioned sub-problems. With $10$ windows the per-window interval is only $360$~s, easy enough that the network reaches a tight local fit before the next handoff; per-window data starvation, i.e., only some windows contain a SRH-2D snapshot, is more than offset by the warm-start IC. One may expect that more windows would eventually hurt accuracy through accumulated handoff errors; the empirical evidence we report here contradicts that expectation up to at least $10$ windows on this case.

  \item FVM teacher distillation is viable and matches direct anchoring at single-network resolution. SR-E, which replaces the external SRH-2D anchor with an in-loop FVM teacher trajectory, achieves $L_2(h) = 7.49{\times}10^{-2}$ and $L_2(|\mathbf{u}|) = 3.56{\times}10^{-2}$ at $t = 3600$~s, effectively indistinguishable from the standard SRH-2D-anchored SR-B baseline ($7.48{\times}10^{-2}$ and $3.20{\times}10^{-2}$, respectively) and clearly worse than the windowed SR-C/D runs. The Savannah River teacher does inherit a small velocity bias from the FVM solver itself: the FVM teacher solution is not perfectly identical to SRH-2D on this real-world reach (cf.\ Fig.~\ref{fig:savannah_setup}c--f), but at single-network resolution that bias is comparable to the residual error already present in SR-B. The FVM teacher variant is therefore especially useful for operational deployments where running a calibrated SRH-2D model is impractical (the in-loop FVM generator is self-contained), and on cases where a calibrated solver is available, the windowed runs (SR-C/D) remain the higher-accuracy choice.

  \item Sparse-only data is insufficient on a real river. SR-F ($200$ sparse velocity measurements at $t = 3600$~s only) achieves $L_2(h) = 0.18$ and $L_2(|\mathbf{u}|) = 0.072$ at the final time, much worse than the dense-anchor SR-B baseline. The transient is more dramatically degraded: at $t = 720$~s the network's depth error is $1.8$~m, i.e., the sparse end-time measurement cannot constrain the path the network takes from a flat-WSE IC to the steady state. On real-world reaches with non-trivial topography and multiple Manning zones, dense multi-snapshot reference data spanning the full time window is necessary.

  \item The FVM-PINN loss is a marginal regulariser at this data density. SR-G (same dense anchor as SR-B but with $\lambda_{\mathrm{fvm-pinn}} = 0$) matches SR-B within statistical noise ($L_2(h) = 0.073$ vs.\ $0.075$; $L_2(|\mathbf{u}|) = 0.032$ vs.\ $0.032$), confirming the BIC-G/H finding from Case 3: at dense data the FVM-PINN loss
    contributes little above the data loss. The FVM-PINN loss's value on real cases is therefore concentrated in the sparse-data regime (where it would be needed if SR-F had been run with $\lambda_{\mathrm{fvm-pinn}} > 0$, or in any operational setting with limited field measurements) and as a structural prior that keeps predictions physically consistent across the wet-dry fringe.
\end{itemize}

\subsection{Scalability Strategies and Their Trade-offs}
\label{sec:strategy_comparison}

\begin{figure}[htp]
  \centering
  \includegraphics[width=0.95\textwidth]{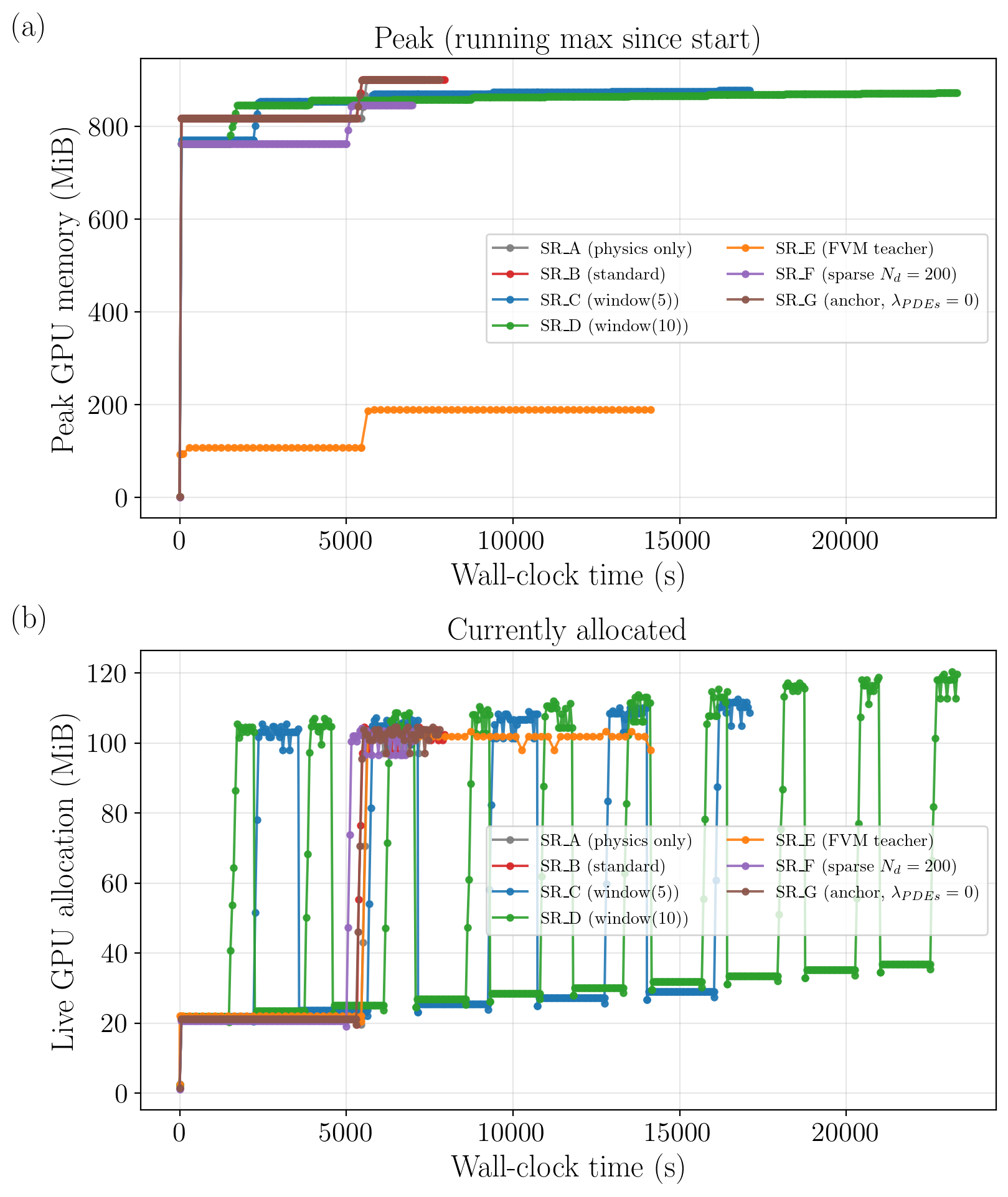}
  \caption{Case 4 GPU memory footprint of the seven Savannah River runs vs.\ wall-clock training time. (a) Running maximum of peak GPU memory since training started. Per-step memory is dominated by the autograd graph of the FVM-PINN loss evaluated over the full mesh at every Adam step. As a result, the strategies that evaluate the FVM residual every step (SR-A, SR-B, SR-F, SR-G) all reach $\sim$$900$~MiB peak. Time windowing (SR-C, SR-D) does not reduce peak memory in any meaningful way and reaches $\sim$$870$~MiB; the time-window strategy's payoff is on the accuracy axis (Table~\ref{tab:savannah}, Fig.~\ref{fig:savannah_l2}), not the memory axis. The only strategy that materially shrinks peak memory is teacher distillation (SR-E $<200$~MiB): the Roe solver runs once up-front to produce cached snapshots, so the gradient computation never has to back-propagate through it. (b) Live GPU allocation, dominated by the network parameters and only weakly dependent on strategy ($\sim$$100$~MiB across runs).}
  \label{fig:savannah_memory}
\end{figure}

The framework implements three scalability strategies for handling larger meshes or longer simulations (Section~\ref{sec:scalability}): time-window decomposition, cell mini-batching with stencil expansion, and gradient checkpointing. The teacher-generated FVM trajectory is a fourth, complementary tool, but in our framework it is framed as a data-guidance variant (Section~\ref{sec:data_guided}) rather than a scalability strategy per se, since its primary effect is to substitute for externally-provided supervision. Cases 3 and 4 exercise the time-window decomposition (SR-C/D) and the FVM teacher data guide (BIC-D, SR-E); mini-batching and gradient checkpointing are only documented in the companion code repository. Here we summarise the qualitative behaviour observed for the time-window strategy and the FVM teacher data guide:
\begin{itemize}
  \item For Case 4, time-window decomposition (Section~\ref{sec:timewindow}) targets the accuracy axis of scalability rather than the memory axis: the per-step autograd graph still spans the entire mesh and the same number of time steps within each window, so the peak GPU footprint of SR-C/SR-D is essentially identical to the single-network SR-B (Fig.~\ref{fig:savannah_memory}, $\sim$$870$~MiB vs.\ $\sim$$900$~MiB). What windowing buys, as Case~4 demonstrates, is a substantial accuracy gain on long-time problems thanks to the IC handoff in equation~\eqref{eq:window_handoff}: each sub-network is trained on a well-conditioned short-horizon sub-problem warm-started from its predecessor. The cost is wall-clock (each window trains a fresh network sequentially), not GPU memory, for Case 4. One scenario that this work did not explore is the use of fewer time steps per window, which would reduce the per-window memory footprint and allow more windows to be used without increasing wall time; this is a promising direction for future work.
  \item FVM teacher data guide (Section~\ref{sec:data_guided}, BIC-D and SR-E) is roughly $3\times$ slower than direct standard training at comparable accuracy on the synthetic block-in-channel case (BIC-D matches the dense-anchor BIC-H), and on the real-world Savannah River case it matches the standard SRH-2D-anchored baseline SR-B almost exactly while remaining behind the windowed runs SR-C/D. It is most useful when the cost of a physics-based solver such as an SRH-2D reference run is prohibitive or not available, but a self-contained FVM teacher fits in memory; it also has a markedly smaller GPU memory footprint than the other runs shown in Fig.~\ref{fig:savannah_memory}, which makes it the lightest-weight option for memory-constrained deployments.
\end{itemize}

\section{Discussion}
\label{sec:discussion}

\subsection{Why physics-only training may fail: a loss-landscape diagnostic}
\label{sec:loss_landscape}

\begin{figure}[htp]
  \centering
  \includegraphics[width=0.95\textwidth]{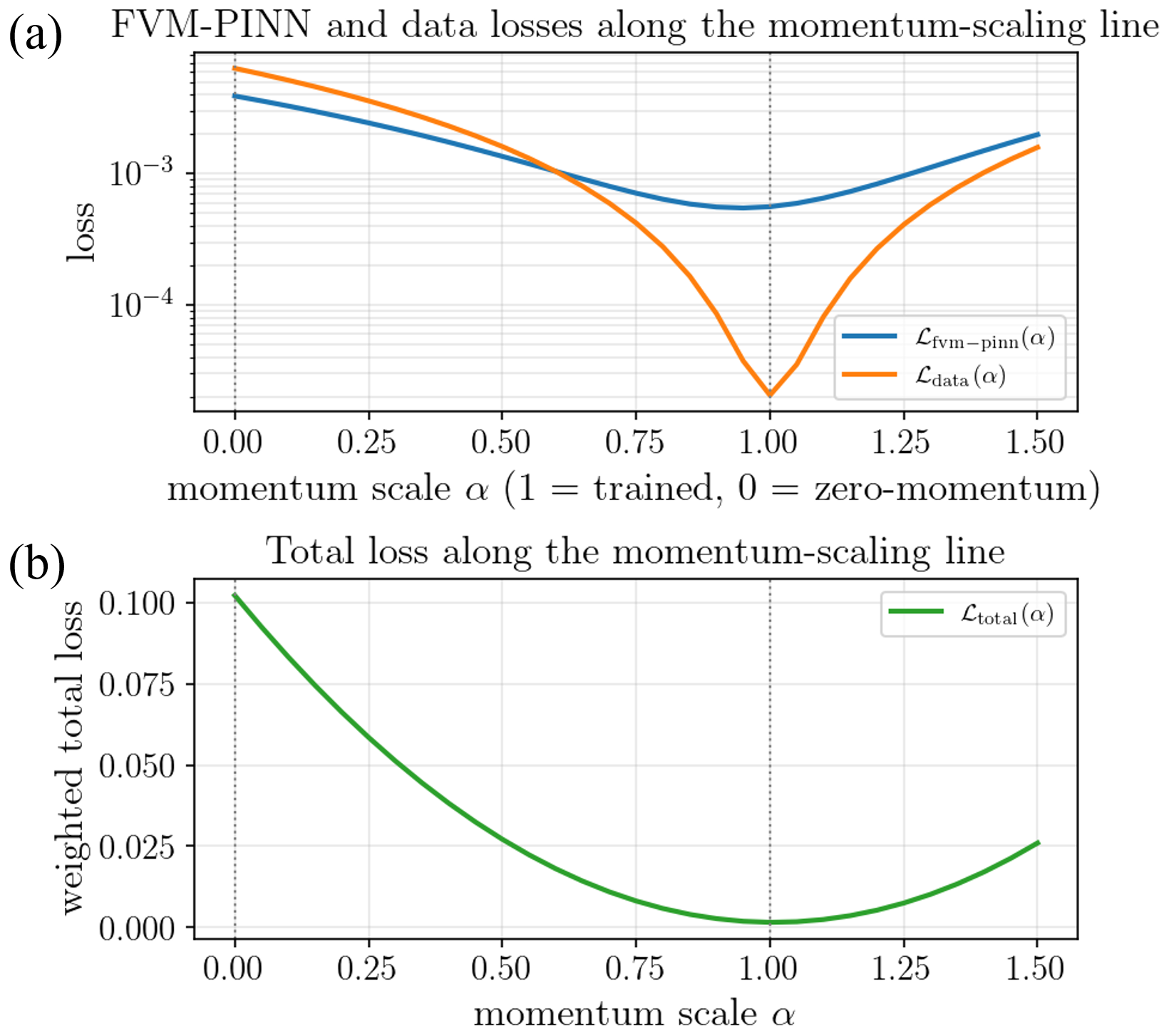}
  \caption{Loss-landscape diagnostic on the BIC-B trained network (block-in-channel): (a) $\mathcal{L}_{\mathrm{fvm-pinn}}(\alpha)$ and $\mathcal{L}_{\mathrm{data}}(\alpha)$ along the momentum-scaling line~\eqref{eq:alpha_perturb}. (b) weighted total loss $\lambda_{\mathrm{fvm-pinn}} \mathcal{L}_{\mathrm{fvm-pinn}} + \lambda_{\mathrm{data}} \mathcal{L}_{\mathrm{data}} + \cdots$. The FVM-PINN loss at $\alpha = 0$ (zero momentum) is only about $7.0\times$ larger than at the trained solution ($\alpha$ = 1) -- a shallow basin that an ordinary first-order optimizer can fall into from a wide range of initialisations. The data loss, in contrast, is about $310\times$ larger at $\alpha$ = 0 than at $\alpha$ = 1, providing the sharply localised attractor that breaks the degeneracy.}
  \label{fig:loss_landscape}
\end{figure}

The failure of BIC-A is straightforward to describe. The network simply collapses to near-uniform depth with near-zero momentum. But it is worth quantifying \emph{why} the FVM-PINN (PDEs) loss permits this state. We make the diagnosis directly visible in Figure~\ref{fig:loss_landscape}. Starting from the trained BIC-B network (which has correctly learned the wake), we apply a one-parameter perturbation
\begin{equation}
  \Q_\theta^{\alpha}(\x, t) = \bigl(\xi_\theta,\, \alpha\,(uh)_\theta,\,
                                    \alpha\,(vh)_\theta\bigr),
\label{eq:alpha_perturb}
\end{equation}
and sweep $\alpha \in [0, 1.5]$, evaluating the FVM-PINN loss $\mathcal{L}_{\mathrm{fvm-pinn}}$ and the data loss $\mathcal{L}_{\mathrm{data}}$ on the same network at each $\alpha$. $\alpha$ is a momentum scaling parameter: $\alpha = 1$ is the trained solution; $\alpha = 0$ is the trivial zero-momentum field with the trained depth (a deliberately constructed ``trivial state'' analogous to the BIC-A failure mode).

The numerical signal is unambiguous. At $\alpha = 0$, $\mathcal{L}_{\mathrm{fvm-pinn}} = 3.91{\times}10^{-3}$ versus $5.60{\times}10^{-4}$ at $\alpha = 1$ with a ratio of about $7.0$. The data loss shows the opposite trend: at $\alpha = 0$, $\mathcal{L}_{\mathrm{data}} = 6.36{\times}10^{-3}$ versus $2.05{\times}10^{-5}$ at $\alpha = 1$ with a ratio of $310$. Adding the data loss therefore multiplies the gradient signal toward $\alpha = 1$ by roughly $44\times$ relative to physics alone. 

This is the mechanistic explanation for the BIC-A failure. A network parameterising low momentum throughout the domain satisfies $d \Q / d t \approx 0$ (the trivial state is quasi-steady) and $\sum_f \hat{\mathbf{F}}\,\ell_f \approx 0$ (zero face fluxes for zero velocity), so the FVM-PINN loss is small everywhere except at the inlet face where the boundary condition imposes a non-zero flow. The boundary loss is supported only on a few faces and is dominated numerically by the residual on $n_c \times n_t \sim 10^5$ interior collocation points. Data supervision at interior points provides the missing gradient signal, which is exactly the role the FVM-PINN loss structurally cannot play on its own.

\subsection{When does the FVM-PINN loss help?}

The BIC-G/H and SR-G ablations decompose the contribution of the FVM-PINN loss itself. Comparing physics-on/off pairs at matched data density gives the following picture:
\begin{itemize}
  \item No data (BIC-A and SR-A): only the FVM-PINN loss acts; the optimizer falls into the
    trivial basin on both the synthetic block-in-channel and the real-world Savannah River cases. The FVM-PINN loss is necessary in the sense that without it the network is not trained at all, but it is not sufficient.
  \item Sparse data ($\sim$$10^2$ points): the FVM-PINN loss reduces $L_2(|\mathbf{u}|)$ by $\sim$$23\%$ (BIC-B $7.67{\times}10^{-3}$ vs.\ BIC-G $9.95{\times}10^{-3}$). This is the regime where field-scale hydrodynamics typically operates: ADCP transects, sparse PIV, and gauge data are fundamentally sparse, and the FVM-PINN loss extends the data signal into unobserved regions.
  \item Dense data (full SRH-2D snapshots): the FVM-PINN loss is essentially neutral (BIC-E vs.\ BIC-H differ at the third significant digit; SR-B vs.\ SR-G at the second). With supervision on every cell at multiple times, the data loss already provides the gradient signal the FVM-PINN loss would supply.
\end{itemize}
Based on all above, the FVM-PINN loss is a regulariser whose marginal value is largest in the sparse-data regime, i.e., the regime where physical priors are most needed.

\section{Limitations and Future Work}
\label{sec:limitations}

\subsection{Limitations}

Several limitations bound the conclusions of this work and define the operating regime in which the proposed framework is currently usable. The per-step autograd graph spans the entire mesh, so peak GPU memory grows linearly with $n_c$. The largest case here (Savannah River, $n_c$ = 1{,}306) fits on a Quadro RTX~4000; substantially larger meshes, e.g., $n_c > 10^6$, would require the cell mini-batching or gradient-checkpointing strategies described in Section~\ref{sec:scalability} but not exercised in this paper. In addition, time-window decomposition trades wall-clock for accuracy: the $N$ = 10 Savannah River run (SR-D) takes about $6.5$ hours on the same hardware. For long simulations or large parameter sweeps, the wall-clock cost is the binding constraint rather than the GPU memory.

All four cases reach steady or quasi-steady states by the end of the training interval. Genuinely transient phenomena such as rapidly propagating flood waves on dry beds or storm-surge inundation with unsteadiness are not examined, and wet-dry-front problems are outside the scope: the softplus on the recovered depth $h$ enforces $h \geq 0$ but the framework is not tested on overland flow with extensive dry zones.

The Savannah River results use SRH-2D output as the reference solution. While SRH-2D is itself calibrated against field measurements, direct validation of FVM-PINN predictions against gauge or ADCP data on the same reach was not attempted here.

\subsection{Future Work}

The differentiable framework presented here opens several natural extensions.

First, the FVM-PINN loss is independent of the surrogate parameterisation: it can be plugged into operator-learning architectures (DeepONet, Fourier neural operators) in place of the strong-form residual, yielding conservation-aware operator surrogates that generalise across boundary conditions and parameter values. This is particularly attractive for ensemble-scale applications such as flood-frequency analysis or reservoir-operation optimisation, where many forward solves are required.

Second, the differentiability of the entire pipeline, from network output through the Roe Riemann solver and back, means that inverse problems are immediately available: Manning's roughness inversion \cite{liu2025uswe}, bathymetry inversion \cite{liu2024bathymetry_inversion}, and inlet-discharge calibration can all be posed as backpropagation through the same loss with the parameter of interest treated as a learnable input.

Third, scaling to large operational meshes ($n_c > 10^6$) remains an open problem. Beyond the cell mini-batching strategy described in Section~\ref{sec:minibatch}, domain decomposition along the lines of XPINN \cite{jagtap2020xpinn} and graph-neural-network spatial encoders are promising directions for further reducing the per-step memory footprint.

Finally, validation against field measurements (gauge time series, satellite altimetry, large-scale PIV) on real river reaches and storm-surge events would test the framework at the data-density regime it is ultimately targeted at, and would clarify how much of the SRH-2D-anchored accuracy reported here transfers to genuinely observation-supervised settings.

\section{Conclusions}
\label{sec:conclusions}

We presented Data-Guided FVM-PINN, a framework that combines a differentiable, well-balanced Roe Riemann finite-volume loss with observational or simulation data guidance for the 2D shallow water equations on unstructured meshes. Our main findings are as follows.

Physics-only FVM-PINN may fail due to rugged loss landscapes. Without data guidance, the optimizer may collapse to a trivial low-momentum state that nearly satisfies cell-wise conservation but bears no resemblance to the true flow. A loss-landscape diagnostic (Section~\ref{sec:loss_landscape}) shows quantitatively why: the FVM-PINN loss at zero momentum is only $\sim$$7\times$ deeper than at the trained solution, a shallow basin that an ordinary first-order optimizer (e.g.\ Adam) falls into; data guidance separates the same two states by $\sim$$310\times$ and breaks the degeneracy.

Data-guided training is the practical remedy, with the FVM-PINN loss contributing most in the sparse-data regime. Just $200$ random velocity measurements drop $L_2(|\mathbf{u}|)$ on the block-in-channel benchmark by $22\times$ ($1.66{\times}10^{-1} \to 7.67{\times}10^{-3}$), and $50$ measurements still deliver a $7\times$ reduction. Decomposing the FVM-PINN loss's contribution by physics-on/off ablation shows it reduces $L_2(|\mathbf{u}|)$ by $\sim$$23\%$ when data is sparse and is essentially neutral when dense reference data is available, a regulariser whose marginal value is largest exactly where physical priors are most needed. Modest measurement noise ($5\%$) erodes some of the data-guidance benefit ($L_2(|\mathbf{u}|)$ rises by $\sim$$50\%$) but does not push the optimizer back into the trivial basin.

Real-world surrogate construction works, with monotonic gains from time-window decomposition. The framework handles a $1$~km Savannah River reach ($1,306$ cells, five Manning zones, $3,600$~s simulation) using SRH-2D anchor data; time-window decomposition reduces $L_2(h)$ monotonically from $0.075$ (single network) to $0.068$ ($5$ windows) to $0.045$ ($10$ windows), driven by progressive IC handoff converting one long-time problem into a chain of well-conditioned sub-problems.

These results recast the FVM-PINN narrative from ``physics-loss is the key idea'' to ``physics-loss and data are complementary, with physics most valuable in the sparse regime and structural priors carrying the bulk of the load when data is dense.'' Because the SWEs underpin a broad range of geophysical fluid applications, the data-guided FVM-PINN methodology and the ablation analysis introduced here are directly transferable to those settings; whether the FVM-PINN loss continues to add measurable value as field measurement densities grow, e.g., from satellite-derived water surface fields, large scale PIV, or remotely sensed coastal inundation, is an open question that our framework can answer.

\section*{Open Research Section}

The FVM-PINN code, training configurations, and example SRH-2D case files supporting the conclusions of this study are publicly available at \url{https://github.com/psu-efd/HydroNet} and archived at Zenodo with DOI \url{https://doi.org/10.5281/zenodo.20099995}. All figures and tables in this manuscript can be reproduced by running the example training scripts on the released configurations and applying the post-processing utilities included with the code release.

\section*{Conflict of Interest disclosure}

The authors declare there are no conflicts of interest for this manuscript.

% ============================================================
\acknowledgments

This work is supported by a seed grant from the Institute of Computational and Data Sciences at the Pennsylvania State University.

% ============================================================
\bibliography{references}

% ============================================================
\appendix

\section{Network Hyperparameters}
\label{app:hyperparams}

\begin{table}[htp]
\centering
\caption{Network and training hyperparameters used in this paper. Cases 1--2 are 1D validation problems; Cases 3--4 are the 2D block-in-channel and Savannah River ablations. Entries marked ``--'' indicate that the hyperparameter is not used by the trainer for that case. Case~2 (BP) is run in two configurations: a Standard trainer without data guide (uses the $\lambda_{\mathrm{fvm-pinn}}/\lambda_{\mathrm{IC}}/\lambda_{\mathrm{BC}}/\lambda_{\mathrm{data}}$ block) and a Teacher trainer with FVM-trajectory distillation (uses the $\lambda_{\mathrm{phys}}/\lambda_{\mathrm{anchor}}$ block); both rows are populated for the BP column. Similarly, BIC-D (Case~3) and SR-E (Case~4) use the Teacher trainer.}
\label{tab:hyperparams}
\small
\begin{tabular}{lcccc}
\toprule
\textbf{Parameter} & \textbf{Case 1 (DB)} & \textbf{Case 2 (BP)}
  & \textbf{Case 3 (BIC)} & \textbf{Case 4 (SR)} \\
\midrule
Hidden dimension       & 64  & 64  & 128 & 128 \\
Number of layers       & 5   & 6   & 6   & 6   \\
Fourier features       & 32  & 32  & 64  & 64  \\
Fourier scale $\sigma$ & 2.0 & 1.0 & 1.0 & 0.5 \\
Activation             & $\tanh$ & $\tanh$ & $\tanh$ & $\tanh$ \\
Residual connections   & No  & Yes & Yes & Yes \\
\midrule
Adam learning rate     & $10^{-3}$ & $10^{-3}$ & $10^{-3}$ & $10^{-3}$ \\
Adam decay factor      & $\times 0.9$ & --   & $\times 0.5$ & $\times 0.5$ \\
Adam decay every       & 1000 & --   & 2000 & 2500 \\
Adam epochs$^{*}$    & 5000 & 3000 & 8000 & 3000--10000 \\
L-BFGS epochs$^{*}$  & 0    & 200  & 500  & 200--500 \\
Time samples per step  & 5    & 36   & 10   & 10 \\
\midrule
$\lambda_{\mathrm{fvm-pinn}}$  & 1.0  & 1.0  & 1.0 (or 0)  & 1.0 (or 0) \\
$\lambda_{\mathrm{IC}}$    & 20.0 & 10.0 & 10.0 & 10.0 \\
$\lambda_{\mathrm{BC}}$    & 5.0  & 30.0 & 30.0 & 30.0 \\
$\lambda_{\mathrm{data}}$  & 0.0  & 0.0  & 10.0 (or 0) & 10.0 \\
\midrule
FVM teacher trajectory $\lambda_{\mathrm{phys}}$   & -- & 0.05 & 0.05 (BIC-D) & 0.05 (SR-E) \\
SRH-2D anchor $\lambda_{\mathrm{anchor}}$ & -- & 0.0  & 0.1 (BIC-D)  & 1.0 (SR-E) \\
\bottomrule
\end{tabular}
\end{table}

The hyperparameters in Table~\ref{tab:hyperparams} reflect a balance between standard PINN practice and per-case tuning, rather than exhaustive grid search. The two-phase Adam + L-BFGS training schedule follows the standard PINN protocol \cite{raissi2019pinn, lu2021deepxde}, and the Fourier-feature embedding \cite{tancik2020fourier_features} mitigates the spectral bias of plain MLPs; its scale $\sigma$ is the only architectural hyperparameter we tune per-case, based on the spatial frequency content of each problem (smaller $\sigma$ for the larger Savannah River domain). The loss weights are fixed by order-of-magnitude scale-balancing of the typical loss-term values: $\lambda_{\mathrm{fvm-pinn}} = 1.0$ is the reference scale, $\lambda_{\mathrm{IC}}$, $\lambda_{\mathrm{BC}}$ and $\lambda_\mathrm{data}$ are adjusted so they provide gradient signals of comparable magnitude. Adaptive loss-weighting schemes can be explored in future work.

\end{document}